%% file: main.tex
\documentclass[transmag]{IEEEtran}
\usepackage{algorithm}
\usepackage{multirow}
\usepackage{makecell}
\usepackage[colorlinks,linkcolor=blue,anchorcolor=blue,citecolor=green]{hyperref}

\usepackage{amsfonts}
\usepackage{booktabs}
\usepackage{color}

\usepackage{subfigure}

\usepackage{times}
\usepackage{epsfig}
\usepackage[cmex10]{amsmath}
\usepackage{amssymb}
\usepackage{multirow}
\usepackage{stfloats}
\usepackage{tabularx}
\usepackage{pifont}
\usepackage{makecell}
\usepackage{threeparttable}
\usepackage{mathtools}
\usepackage{xcolor}
\usepackage{graphicx}
\usepackage{caption}
\usepackage{algorithmic}

\usepackage{url}
\begin{document}

\title{Horizontal-to-Vertical Video Conversion}

\author{
Tun~Zhu,~Daoxin~Zhang,~Yao~Hu,~Tianran~Wang,~Xiaolong~Jiang,\\  Jianke~Zhu~\IEEEmembership{Senior~Member,~IEEE},~Jiawei~Li
\IEEEcompsocitemizethanks{
\IEEEcompsocthanksitem Tun~Zhu and Jianke Zhu are with the College of Computer Science, Zhejiang University, Hangzhou, China, 310027. Jianke Zhu is also with the Alibaba-Zhejiang University Joint Research Institute of Frontier Technologies, and Zhejiang Provincial Key Laboratory of Service Robot, College of Computer Science, Zhejiang University. \protect\\
E-mail: \{ianzhu, jkzhu\}@zju.edu.cn.
\IEEEcompsocthanksitem Daoxin~Zhang, Yao~Hu, Tianran~Wang, Xiaolong~Jiang and Jiawei~Li are with company of Alibaba, Youku Cognitive and Intelligent Lab.\protect\\
E-mail: \{daoxin.zdx, yaoohu, steve.wtr, xainglu.jxl, mingong.ljw\}@alibaba-inc.com.
\IEEEcompsocthanksitem Jianke Zhu is the Corresponding Author. The work is supported by National Natural Science Foundation of China under Grants (61831015). Our implementation is available at https://github.com/JackieZhangdx/H2V.
}
\thanks{}}

\maketitle

\begin{abstract}
At this blooming age of social media and mobile platform, mass consumers are migrating from horizontal video to vertical contents delivered on hand-held devices. Accordingly, revitalizing the exposure of horizontal video becomes vital and urgent, which is hereby tackled by our automated horizontal-to-vertical (abbreviated as {\it \textbf{H2V}}) video conversion framework. Essentially, the {\it \textbf{H2V}} framework performs subject-preserving video cropping instantiated in the proposed Rank-SS module. Rank-SS incorporates object detection to discover the candidate subjects, from which we select the primary subject-to-preserve leveraging location, appearance, and salient cues in a convolutional neural network. In addition to converting horizontal videos vertically by cropping around the selected subject, automatic shot detection and multi-object tracking are integrated into the {\it \textbf{H2V}} framework to accommodate long and complex videos. To develop {\it \textbf{H2V}} systems, we collect an {\it \textbf{H2V-142K}} dataset containing 125 videos (132K frames) and 9,500 cover images annotated with primary subject bounding boxes. On {\it \textbf{H2V-142K}} and public object detection datasets, our method demonstrates promising results on the subject selection comparing to the related solutions. Furthermore, our {\it \textbf{H2V}} framework is industrially deployed hosting millions of daily active users and exhibits favorable {\it \textbf{H2V}} conversion performance. By making this dataset as well as our approach publicly available, we wish to pave the way for more horizontal-to-vertical video conversion research. Our collected {\it \textbf{H2V-142K}} dataset is available at \href{https://tianchi.aliyun.com/dataset/dataDetail?dataId=93339}{{\it \textbf{H2V-142K website}}}. 
\end{abstract}


\IEEEpeerreviewmaketitle

\input{intro}
\input{dataset}

\input{framework}
\input{subselect}
\input{experiment}
\input{conclusion}

\ifCLASSOPTIONcaptionsoff
  \newpage
\fi

\bibliographystyle{abbrv}
\bibliography{ref}

\begin{IEEEbiography}[{\includegraphics[width=1in,height=1.25in,clip,keepaspectratio]{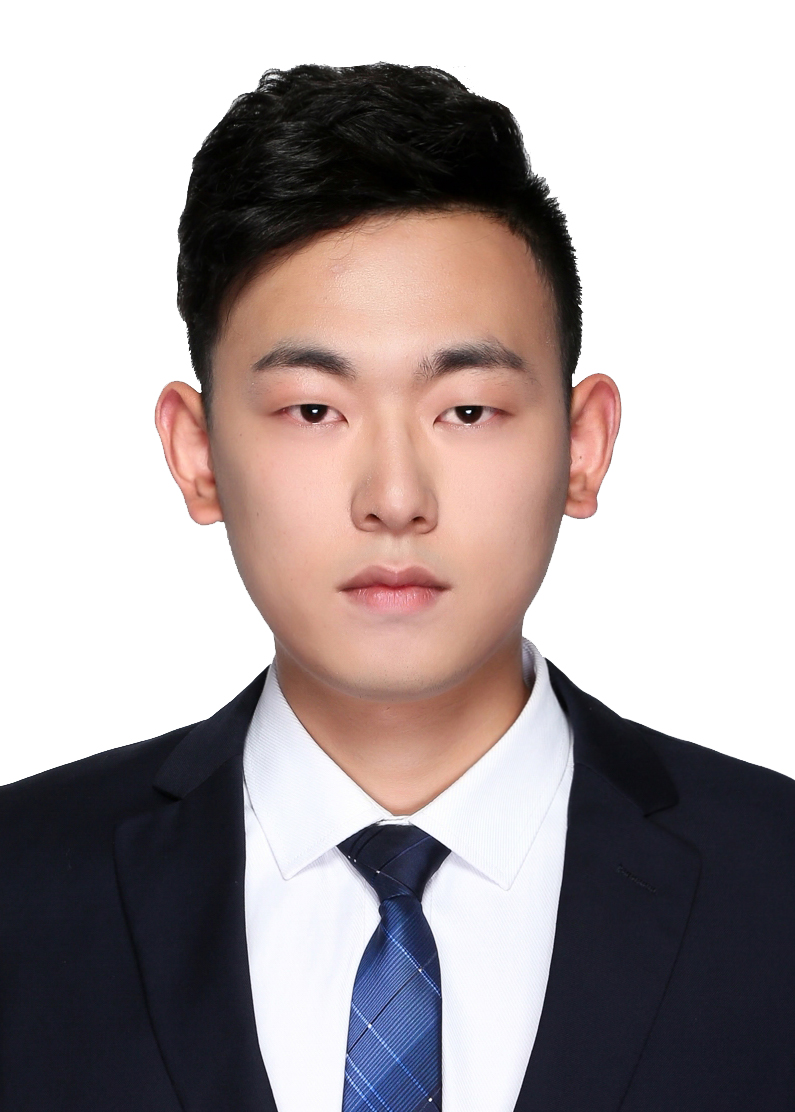}}]{Tun~Zhu} received his bachelor's degree in Computer Science and Technology from University of Electronic Science and Technology of China in 2018. He is currently a Master candidate in the College of Computer Science and Technology, Zhejiang University of China. His research interests include machine learning and computer vision.
\end{IEEEbiography}

\begin{IEEEbiography}[{\includegraphics[width=1in,height=1.25in,clip,keepaspectratio]{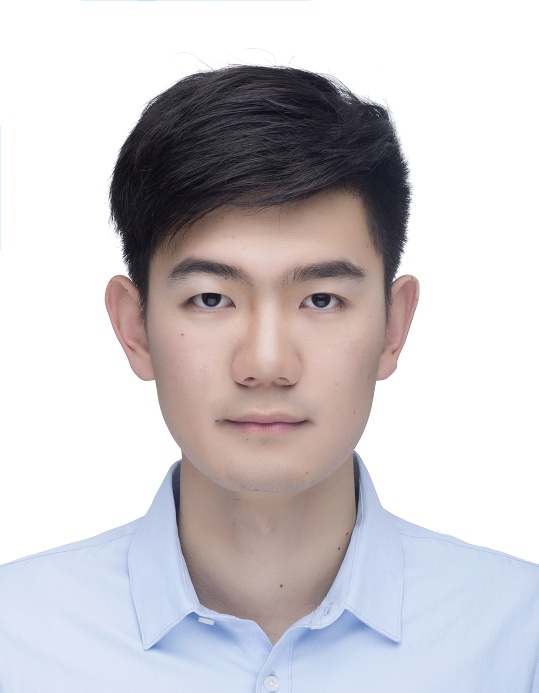}}]{Daoxin~Zhang} received his bachelor’s and master’s degrees in Mathematics and Computer Science respectively both from Zhejiang University in 2016 and 2019. He is currently an Algorithm Engineer II in Alibaba Digital Media\&Entertainment Group. His research interests include multimedia, machine learning and computer vision.
\end{IEEEbiography}

\begin{IEEEbiography}[{\includegraphics[width=1in,height=1.25in,clip,keepaspectratio]{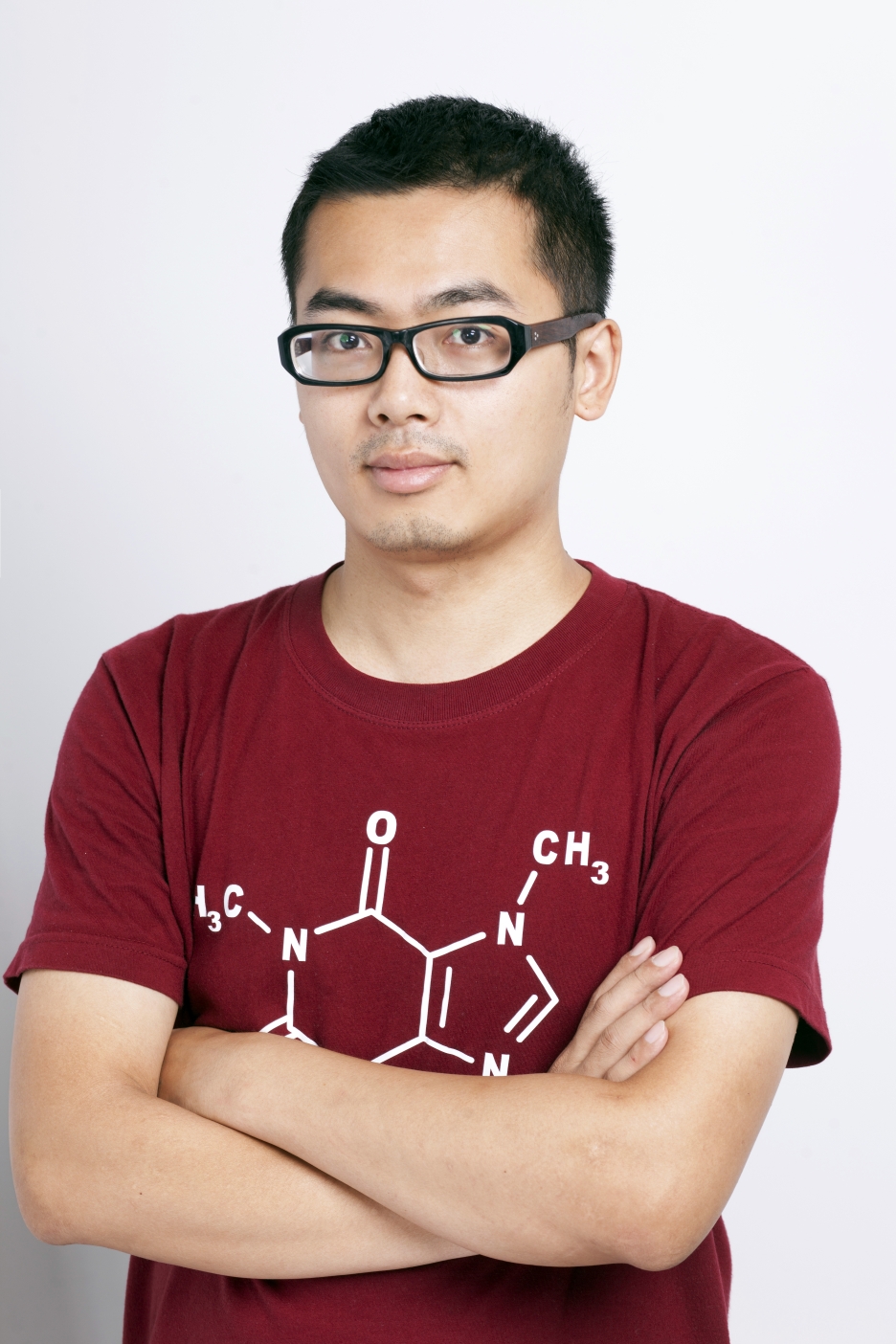}}]{Yao~Hu} is a Senior Staff Algorithm Engineer of Alibaba, now leads the Artificial Intelligence Department of Alibaba Digital Media\&Entertainment Group. He receives the PhD degree in machine learning from Zhejiang University in 2015. Before joining Alibaba, he leads the machine learning team working on DiDi’s travel business and autonomous car. His research interests include machine learning, online learning theory and computer vision. He has published more than 40 papers in TPAMI, SIGKDD, ICML, CVPR, ICCV, IJCAI, and AAAI.
\end{IEEEbiography}

\begin{IEEEbiography}[{\includegraphics[width=1in,height=1.25in,clip,keepaspectratio]{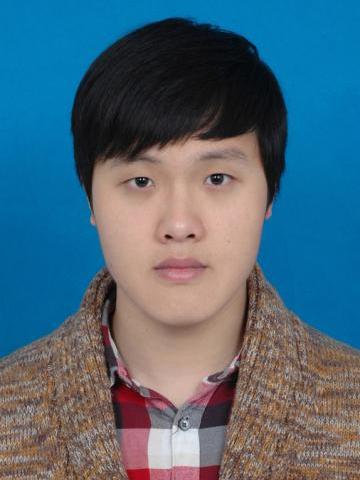}}]{Tianran~Wang} received his bachelor's degree in Computer Science and Technology from Hangzhou Dianzi University in 2018 and master's degree in Software Engineering from Zhejiang University in 2020. He is currently an Algorithm Engineer of Alibaba. His research interests include machine learning and computer vision.
\end{IEEEbiography}

\begin{IEEEbiography}[{\includegraphics[width=1in,height=1.25in,clip,keepaspectratio]{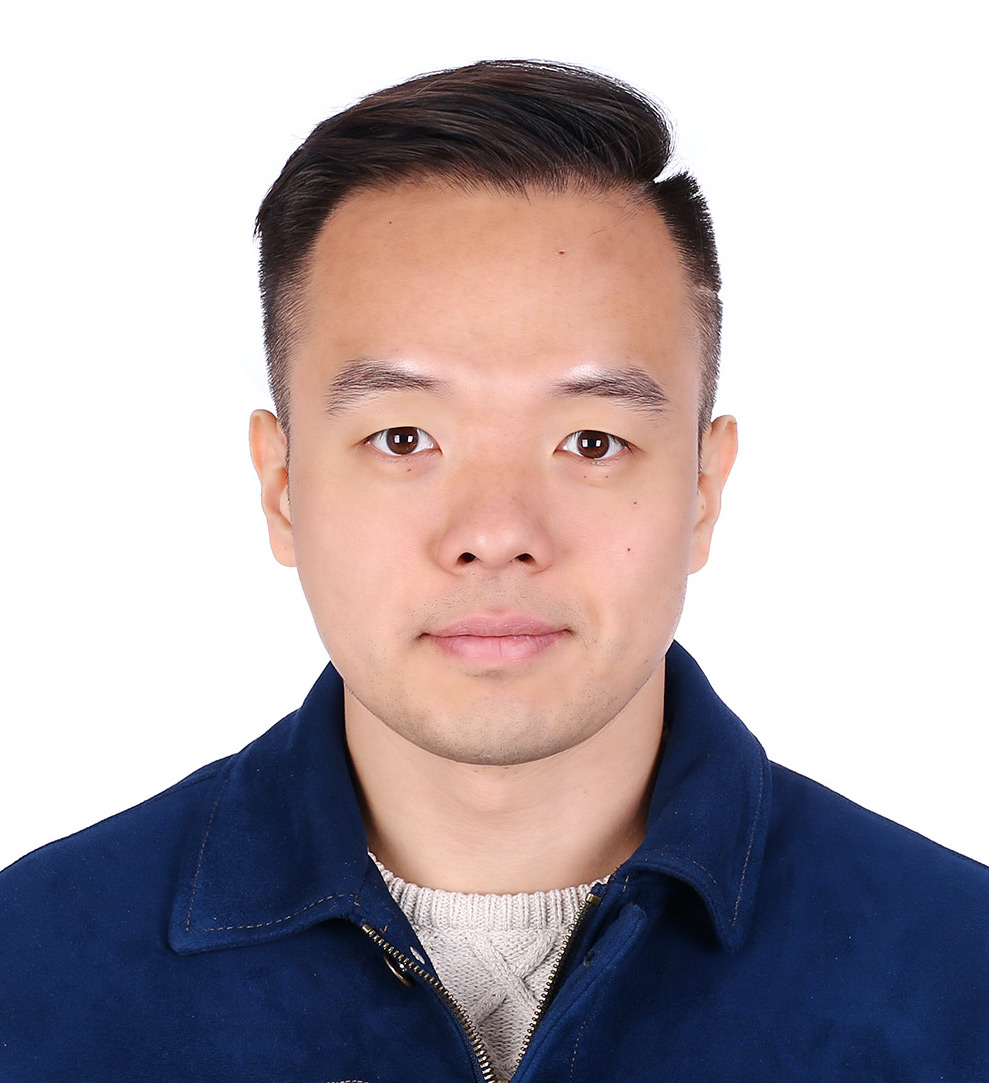}}]{Xiaolong~Jiang} received the bachelor degree in electrical engineering from Beihang University in 2012, and acquired his master degree in computer engineering from Columbia University in 2014. He achieved his PhD degree in electrical engineering from Beihang University in 2019, during which he joined CRCV in University of Central Florida as a visiting scholar. He was a research intern at IDL lab in Baidu in 2018, and now is a researcher in Alibaba Digital Media\&Entertainment Group.
\end{IEEEbiography}

\begin{IEEEbiography}[{\includegraphics[width=1in,height=1.25in,clip,keepaspectratio]{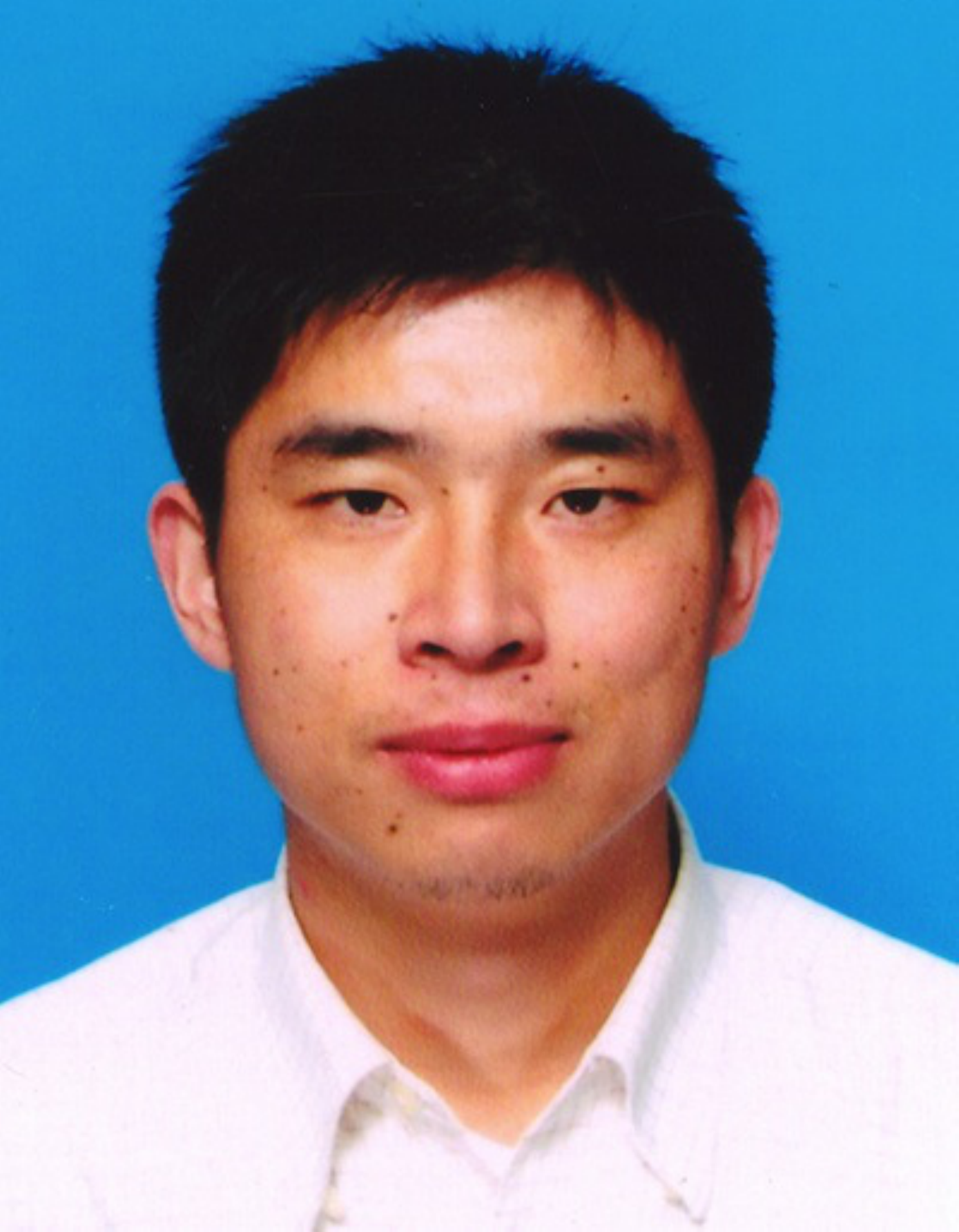}}]{Jianke Zhu} received the master’s degree from  University of Macau in Electrical and Electronics Engineering, and the PhD degree in computer science and engineering from The Chinese University of Hong Kong, Hong Kong in 2008. He held a post-doctoral position at the BIWI Computer Vision Laboratory, ETH Zürich, Switzerland. He is currently a Professor with the College of Computer Science, Zhejiang University, Hangzhou, China. His research interests include computer vision and multimedia information retrieval. He is a Senior member of the IEEE.
 \end{IEEEbiography}

\begin{IEEEbiography}[{\includegraphics[width=1in,height=1.25in,clip,keepaspectratio]{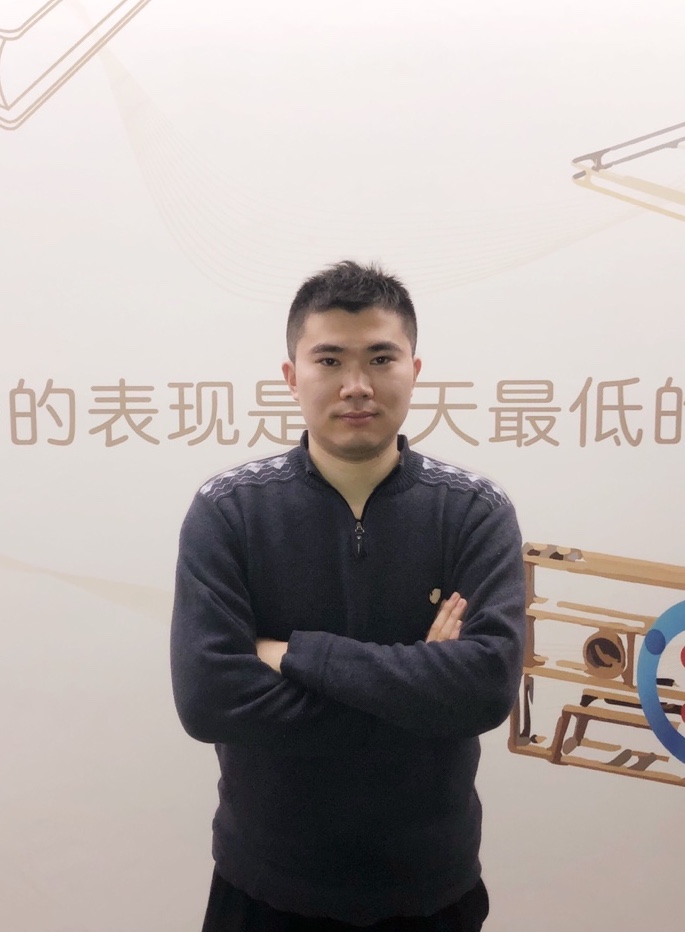}}]{Jiawei~Li} received his bachelor's degree in Computer Science from China Agricultural University in 2012. He is currently a Staff Algorithm Engineer of Alibaba. His research interests include machine learning and recommender system.
 \end{IEEEbiography}
 

%\vfill


\end{document}

%% file: intro.tex
\section{Introduction}
Vertical videos are created for viewing in portrait mode on hand-held devices, which are opposite from traditional horizontal formats popularized on big screens. With the unprecedented growth of social media platforms, such as TikTok, Instagram, and Youku, {\it etc.}, vertical videos take over the focus of mass video consumers, leaving abundant horizontal contents less exposed. To reintegrate their exposure, the horizontal videos have been converted vertically with manual processing and cropping, which is prohibitively labor-intensive and time-consuming. Accordingly, there is an imperative need for fully-automated horizontal-to-vertical video conversion solution.

\begin{figure}[ht]
  \centering
  \includegraphics[width=0.85\linewidth]{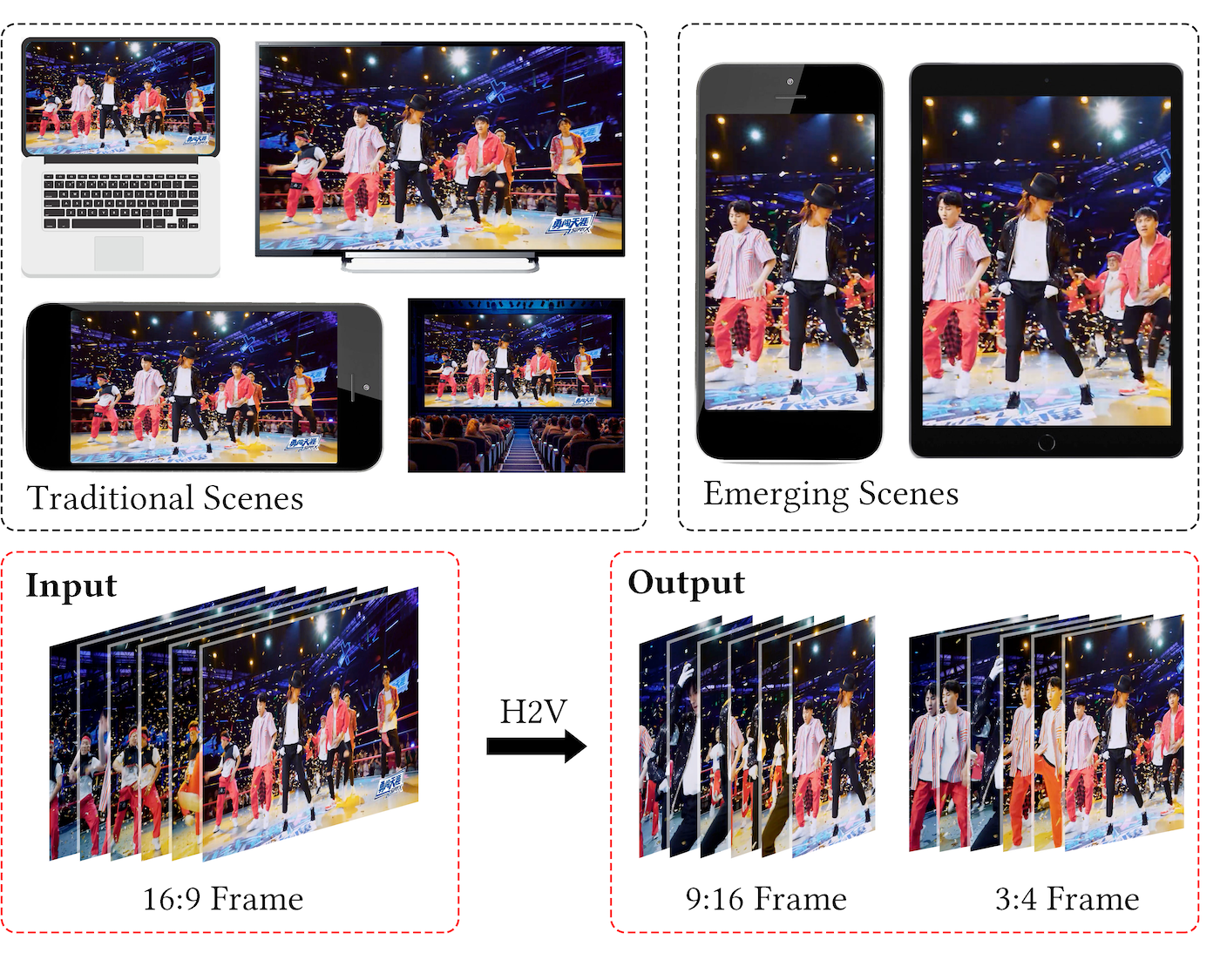}
  \setlength{\belowcaptionskip}{-0.3cm}  
  \caption{Traditional video consumption scenarios such as Films, TV and stream media prefer horizontal frame ({\it e.g.} 16:9 frame). With the growth of emerging scenes like portable devices, vertical video content is in huge demand while lacking professional producers. Our {\it H2V} framework manages to generate vertical videos from high-quality horizontal video inventory.}
  \label{fig:motivation}
\end{figure}

Nonetheless, automated horizontal-to-vertical video conversion is an uncharted territory, of which the key challenge is subject preservation, {\it i.e.} keeping the main subject (mostly human) stable in the scene through the information-losing video cropping. As illustrated in Fig.~\ref{fig:motivation}, conversion solutions implement subject preservation by cropping horizontal sources around the primary subject for producing vertical outputs. To achieve subject-preserving conversion, one needs to develop a fully automated pipeline assembling shot boundary detection, subject selection, subject tracking, and video cropping components. Subject selection is of the most cardinal importance. 

\begin{figure}[htbp]
\centering
\subfigure[Subject Selection (Ours)]{
\includegraphics[width=0.45\linewidth]{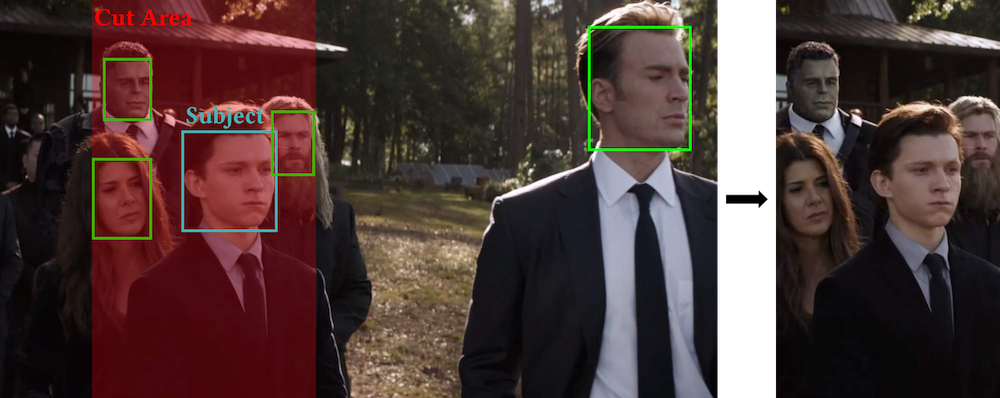}
%\caption{fig1}
}
\subfigure[Object Detection]{
\includegraphics[width=0.45\linewidth]{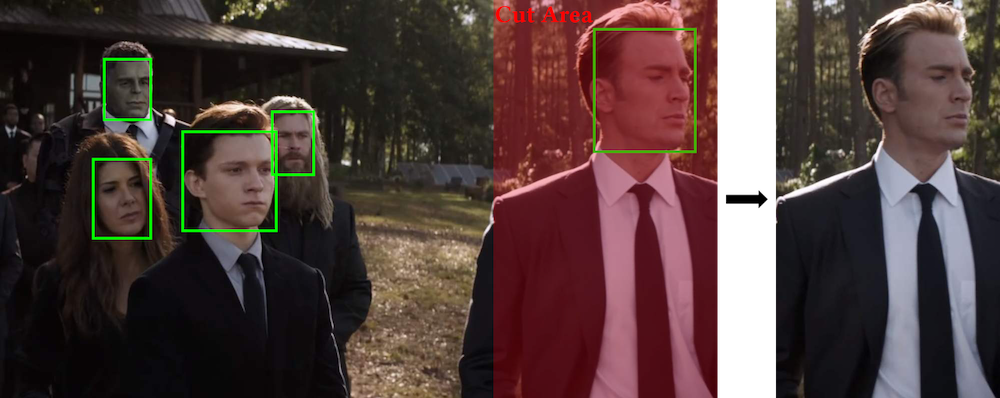}
}
\subfigure[Salient Object Detection]{
\includegraphics[width=0.45\linewidth]{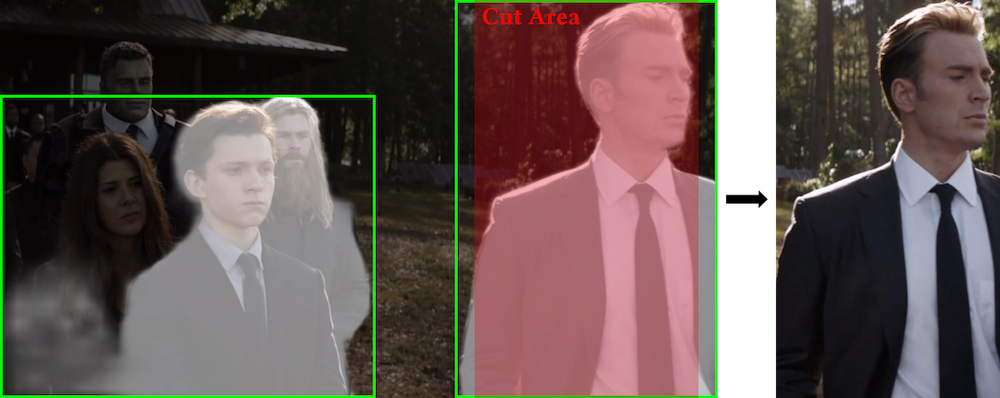}
}
\subfigure[Fixation Prediction]{
\includegraphics[width=0.45\linewidth]{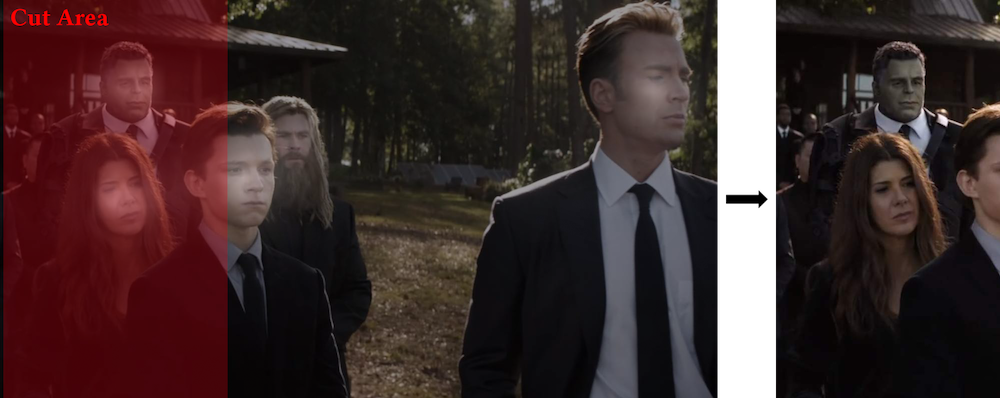}
}
\caption{Comparing the results of our approach with the traditional object detection approach and saliency approaches. The results are illustrated as red cutting area and right images simultaneously. (a) Our method outputs subject localization. (b) Object detection only locates objects with a specific category. (c) Salient object detection outputs binary masks without aware of object instances. (d) Fixation prediction computes the most eye-catching areas ignoring boundary and object instances similarly.}
\setlength{\belowcaptionskip}{-0.3cm}  
\label{fig:task_compare}
\end{figure}

Achieving subject selection in horizontal-to-vertical conversion is complicated for two reasons. Firstly, the primary subject in a video shifts constantly from shot-to-shot, therefore accurate shot boundary detection is indispensable as pre-processing. Secondly, in most cases, one has to select the primary subject out of numerous foreground distractors within each frame. To overcome this challenge, Salient Object Detection (SOD)~\cite{borji2019saliency,wang2019salient} and Fixation Prediction (FP)~\cite{wang2019revisiting} have been practiced. SOD performs the pixel-level foreground-background binary classification to discover all objects but fails to discriminate the primary subject from other candidates. For FP, although being able to find the primary subject by imitating human visual system, yet it can only provide point-like fixation response thus fail to obtain the subject in its entirety. A comparison between these strategies and our solution is shown in Fig.~\ref{fig:task_compare}.

In this work, we propose the {\it H2V} framework as the first automatic horizontal-to-vertical video conversion system, which tackles the subject-preserving video cropping effectively and compactly. As shown in Fig.~\ref{fig:overview}, the {\it H2V} framework first incorporates shot boundary detection to separate horizontal input into disjointed shots, where each one contains its own set of subjects. Within each shot, the primary subject is selected at the first frame with the proposed {\it Rank Sub-Select} (Rank-SS) module. As depicted in Fig.~\ref{fig:subselect}, Rank-SS employs a convolutional architecture and integrates human detection, saliency detection, traditional as well as deep appearance features to discover and select the primary subject. In the following frames within the shot, the primary subject is propagated via the subject tracking module. For the development of Rank-SS module, we start by investigating naive and deep regression approaches emphasizing on location priors, which report inadequate performance. Then, we extend the regression into a ranking formulation for primary subject selection, where object-pair relation is taken into consideration and delivers favorable performance. The detailed insights with respect to this extension are elaborated in Section \ref{sec:rankss}. 

\begin{figure*}[!t]
  \centering
  \includegraphics[width=0.9\linewidth]{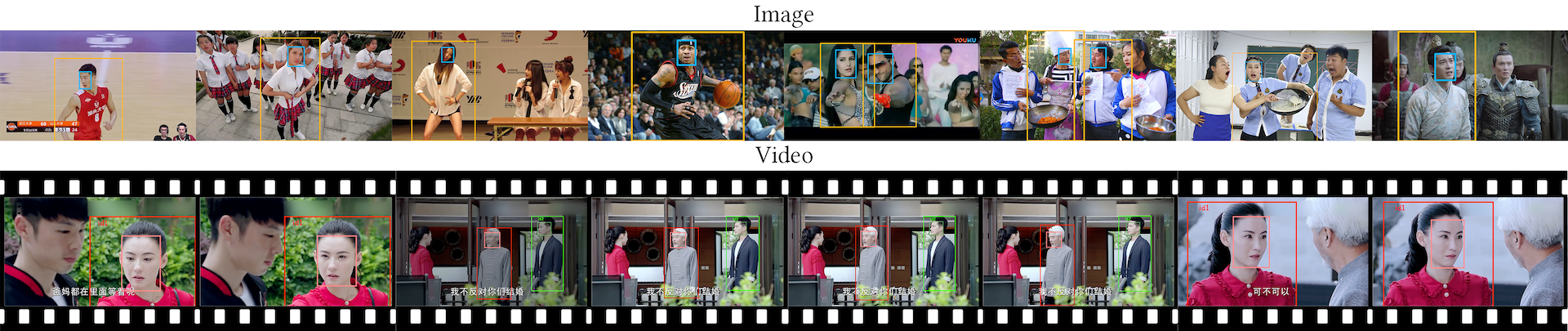}
  \caption{Sample images of {\it H2V-142K} Dataset. We annotate subjects (mostly one) consisting of the face and body.}
  \label{fig:dataset}
\end{figure*}

To build and evaluate our {\it H2V} framework, as well as to encourage further researches for horizontal-to-vertical video conversion, we collect and publicize a large-scale dataset named as {\it H2V-142K}.  {\it H2V-142K} dataset contains 132K frames from 125 video sequences, which were carefully labeled by human annotators with bounding boxes denoting the face and torso of the primary subject. As shown in Fig.~\ref{fig:dataset}, this dataset covers rich horizontal contents, including TV series, variety shows, and user-made videos. Besides, another 9,500 video cover images with heavy distractors are provided, promoting more robust primary subject selection. Please refer to Section \ref{sec:dataset} for the detailed statistics of the dataset. On top of the {\it H2V-142K} dataset, the publicized detection dataset such as ASR, and extensive user feedback data were collected from Youku website. We have conducted comprehensive experiments, where our {\it H2V} framework exhibits favorable qualitative performance, both quantitatively and qualitatively.

In summary, we highlight three main contributions in this paper: 1) we propose the {\it H2V} framework as the first unified solution to settle horizontal-to-vertical video conversion, which has been successfully commercialized at a web-scale; 2) the {\it Sub-Select} module is designed in the {\it H2V} framework, among other well-performing components, which integrates rich visual cues to select the primary subject in a ranking manner; 3) we construct and publicize the {\it H2V-142K} dataset with 125 fully-annotated videos (more than 142k images), hoping to pave the way for future endeavors in the field of horizontal-to-vertical video conversion.

\section{Related Work}

This section summarizes related works of the task, which is treated as a content-aware video cropping problem. We firstly survey traditional cropping methods and illustrate the limitation in this specific issue. Research on salient object detection and fixation prediction is introduced that closely related to subject discovery and selection. Moreover, instances localization and ranking are required to get the most eye-catching subject. 

\subsection{Image Cropping}

Image cropping is a significant technique for improving the visual quality of raw images. Early methods leverage the practical experience from photographical experts to solve this problem, e.g rule of central, rule of thirds, rule of grid. With the development of deep learning and arising of large-scale aesthetic datasets, like AVA\cite{murray2012ava}, AADB\cite{kong2016photo}, researchers recently solve this task in a data-driven manner and have made great progress in this area. Modern DL-based image cropping methods could be categorized into two streams: structure-based and aesthetic-based. 1) \textbf{Structure-based methods} \cite{sun2013scale,chen2016automatic,wang2016stereoscopic} focus on preserving the most important or salient part after cropping. Attention-mechanism or salient detection ideas are usually applied in these methods. 2) \textbf{Aesthetics-based methods} \cite{hosu2019effective,yan2013learning,zeng2019reliable,fang2014automatic} improve the cropping results by increasing aesthetic quality, local factors are highly considered, which are in favor of preserving visually attractive parts. Also, there are methods that combine both global structure and local aesthetic. \cite{wang2018deep} models image cropping in a determining-adjusting framework, which first uses attention-aware determining, and then applies aesthetic-based adjusting network. \cite{tu2020image} designs the composition-aware and saliency-aware blocks to select more reasonable cropping. 

Image cropping technique is also extended to video retargeting task, which is the process of adapting a video from one screen resolution to another to fit different displays~\cite{rubinstein2010comparative}. Traditional methods argue that important objects in the images like faces or text should be preserved and these algorithms are called \textbf{content-aware algorithms}. Differently from single image, temporal stability is highly relevant in frames and video flickers should try to be avoided. Most of approaches optimize the cropping process shot by shot, and then apply a path generation algorithm to obtain a smooth cropping result \cite{liu2016towards,rubinstein2010comparative,vaquero2010survey}. Although having made great efforts, to the best of our knowledge, there are limited methods that are able to handle complex scenes stably such as multiple distractive objects.

\subsection{Salient Object Detection}
Salient Object Detection (SOD) has a long history date back to Itti et al's work~\cite{itti1998model}. The majority of SOD~\cite{borji2019saliency,wang2019salient} methods is designed to detect pixels that belong to the salient objects without knowing the individual instances, which is commonly treated as a pixel-wise binary classification problem. Traditional heuristic SOD research experienced changes from pixel-based methods~\cite{itti1998model,ma2003contrast}, patch-based approaches~\cite{liu2010learning,achanta2008salient} to region-based methods~\cite{cheng2014global,perazzi2012saliency}. Recently, deep learning-based methods dominantly lead the state-of-the-art advances in SOD including Multi-Layer Perceptron-based method~\cite{zhang2016unconstrained,zhao2015saliency}, Fully Convolutional Networks~\cite{liu2018picanet,zhang2019training,wu2019cascaded} and Capsule-based approach~\cite{liu2019employing,qi2019multi}.

Video salient object detection is similar to image SOD problem discussed above. How to encode motion saliency between frames is the central issue for video SOD. The bottom-up strategy is the common practice for heuristic methods employing background removal~\cite{koh2017primary}, points tracking and clustering~\cite{grundmann2010efficient,ochs2011object}, object proposal ranking~\cite{faktor2014video,xiao2016track} to tackle this problem. Based on deep leaning model, motion encoding is achieved by optical flow~\cite{jain2017fusionseg,cheng2017segflow,li2018flow} or recurrent neural network~\cite{li2018flow,tokmakov2017learning,song2018pyramid}. In addition, co-saliency estimation~\cite{jiang2019unified,li2019co} searches for the common salient object regions contained in an image set.

Fixation Prediction (FP)~\cite{wang2019revisiting} is another closely related area investigating the human visual system's attention behavior. Prevalent datasets such as DHF1K~\cite{wang2019revisiting} record participants' eye movement and save it as a fixation map. From a task perspective, fixation prediction only calculates the fixation points or small areas rather than inferring the primary salient objects like SOD. Thus, FP models~\cite{jiang2018deepvs,leifman2017learning,gorji2018going} care about neither object contour nor object instance.

\subsection{Salient Instance Ranking}
Both SOD and FP methods demand expensive pixel-level annotation and generalize poorly in complex scenes while inadequately being able to distinguish multiple objects. Therefore, general Salient Object Subitizing~\cite{zhang2016unconstrained,li2017instance} methods have been proposed to achieve this. For known object categories, object detection~\cite{zhang2019freeanchor,li2019dsfd} is a more accurate solution. In our {\it H2V} framework, we employ a detector for subject discovery as the target subject is human. 

To solve the salient ranking problem, Li et al.~\cite{li2014secrets} found that a strong correlation between fixations and salient object exists. Similarly, Wang et al. proposed ranking video SOD~\cite{wang2019ranking} with ranking saliency module leveraging FP and SOD features, which presented promising results. However, the fixation data is relatively difficult to label, which limits its application. Another solution introduced by Amirul et al.~\cite{amirul2018revisiting} is a hierarchical representation of relative saliency and stage-wise refinement. In their Salient Object Subitizing dataset, prominent objects are asked to label. Furthermore, relative rank scores are computed by averaging the degree of saliency within the instance mask. We summarize it as Ranking by Global Average Pooling (RGAP) scores~\cite{amirul2018revisiting,wang2019ranking}. As a baseline of the subject selection problem, RGAP-based models N-SS and D-SS (described in Section \ref{sec:ndss}) fail to rank the hard cases with strong spatial characteristics such as side face. Our proposed RCNN-based~\cite{ren2015faster,girshick2015fast} Rank-SS model leverages spatial features to achieve the better region-based ranking.

%% file: dataset.tex
\section{The {\it H2V-142K} Dataset}
\label{sec:dataset}
\subsection{Dataset Overview}
The under-development of {\it H2V} conversion is partially attributed to the lack of available data, to which we collect a large-scale {\it H2V-142K} dataset consists of image set and video set. Regarding the train/test split, we randomly select 600 video covers from the image set and all videos in the video set for testing. The remaining images are used for training. 

The ground-truth of this dataset is divided into two parts. For the subject selection task, our dataset provides the subjects' bounding boxes, including face and body. For the {\it H2V} conversion task, the vertical version ground-truth is the maximum cropping rectangles with the size 9:16, which share the same central horizontal coordinate with subjects' bounding boxes (no pixels tailored vertically). Visualization of sample images and annotations is shown in Fig. \ref{fig:dataset}. 

\begin{table}[htbp]
	%\begin{threeparttable}
	\begin{center}
		%\resizebox{0.5\textwidth}{!}{
		\begin{tabular}{l|c|c c}
			\hline
			Type & Amount& Avg \#Subject& Avg \#Person \\ 
			\hline
			\hline
			Images &   9,500 &   1.1417 & 4.2820  \\
			\hline
			Videos & 125(132k frames)  & 1.0049 & 1.8272 \\
			\hline
		\end{tabular}
		%}
	\end{center}
	% \vspace{-0.1in}
	% \captionsetup{singlelinecheck=off, skip=4pt}
	\setlength{\belowcaptionskip}{-0.3cm}   %调整图片标题与下文距离
	\caption{Detailed statistics of {\it H2V-142K} Dataset.}
% 	\caption{Detailed statistics of {\it H2V-142K} Dataset. The image subset is more complicated than the video subset in that the images in the image subset contain more human distractors than the frames of the video in the video subset.}
	\label{table:dataset_stat}
	%\vspace{-0.15in}
	% \end{threeparttable}
\end{table}

\begin{figure}[ht]
  \centering
  \includegraphics[width=0.9\linewidth]{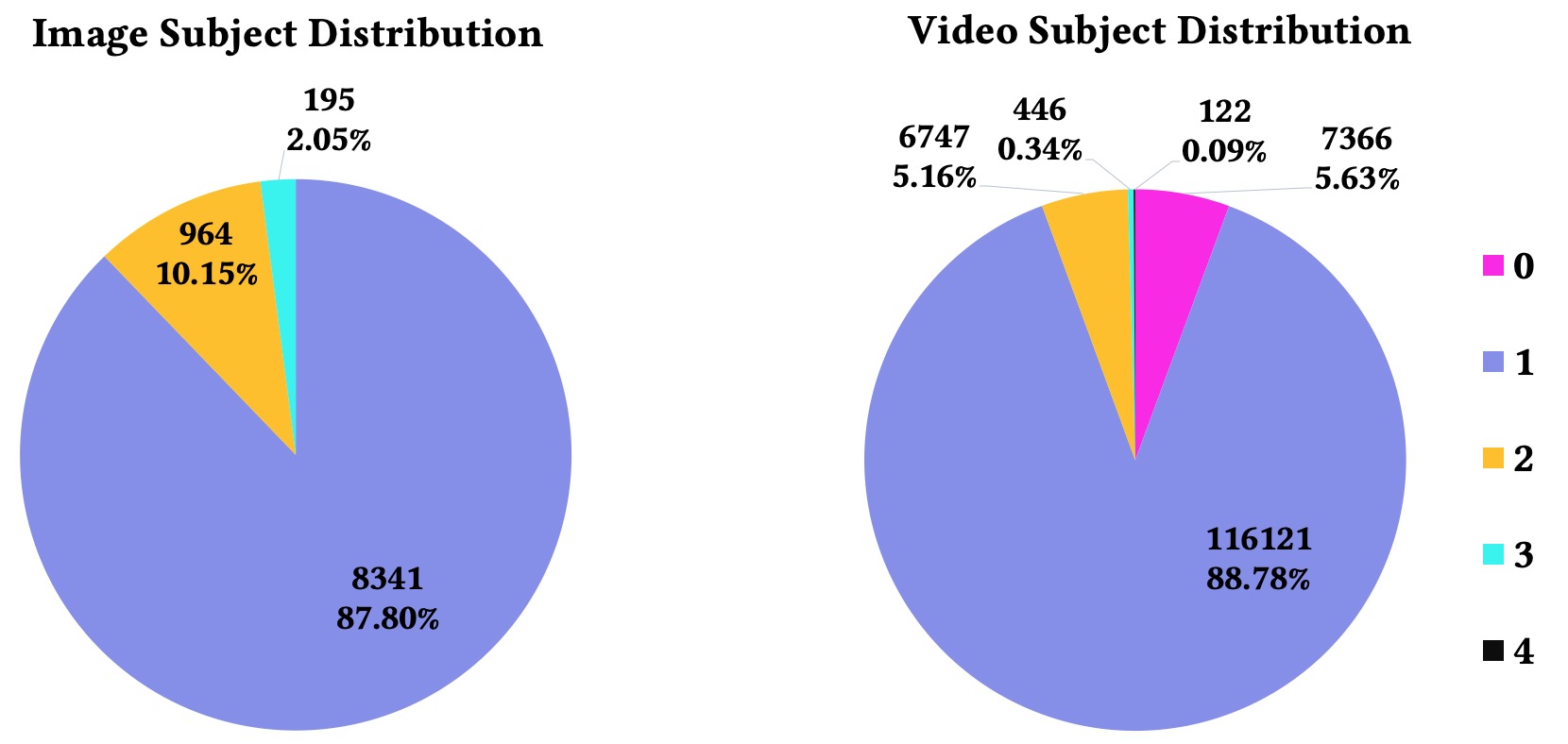}
  \caption{Distribution of subject number in our {\it H2V-142K} Dataset.}
  \label{fig:sub_num}
\end{figure}

\textbf{Analysis.} The {\it H2V-142K} dataset contains 125 videos (132K frames) and 9.5K images. The detailed statistics are shown in Table~\ref{table:dataset_stat}. The video subset covers more general scenes with fewer subjects, while the image subset better examines the performance of subject selection with heavier distractors. Therefore, the average number of person and the total number of subjects are much higher than video subset. The distribution of subjects in two subsets is presented in Fig.~\ref{fig:sub_num}. The majority of images or frames have only one subject. When multiple subjects appear in ground-truth, the result matching any one of them will be regarded as the correct prediction. 

\subsection{Data Collection and Annotation}
The {\it H2V-142K} dataset is collected from Youku video-sharing website and carefully annotated by human annotators following the {\it Sub-Select} criteria, as explained in Section \ref{sec:criteria}. The dataset focuses on the most prevalent scenes, i.e. character-centric scenes. 

\textbf{Video Subset.} We collect videos with the diversified contents and subject scales. Each video is firstly segmented into disjointed shots using TransNet~\cite{souvcek2019transnet}. Within each shot, a group of three annotators, separately annotate every frame with the primary subjects using pairs of the face and torso bounding boxes after watching all frames. The ground-truth bounding boxes are determined by cross-validating over annotators by thresholding on box Intersection-of-Union (IoU). Should a disagreement occur on one frame, it will be carefully reviewed and determined by a group of new annotators. Finally, the bounding boxes are smoothed temporally by Kalman Filter~\cite{Zarchan2001FundamentalsOK}. 

\textbf{Image Subset.} For the majority of instances, only one subject is annotated. In the meantime, co-subjects appear more frequently in the image subset because of the pre-filter during data preparation. Concretely, a human detector FreeAnchor~\cite{zhang2019freeanchor} pre-trained on the COCO dataset~\cite{Lin2014MicrosoftCC} is applied to collect images containing more than three (at least two) human candidates, and the ground-truth subject is annotated in the same manner applied in the video subset. Besides, non-subjects (distractors) are required to rank by the same criterion as complementary labels. When crowds appear, annotators are asked to rank the top six non-subjects, disregarding others to generate hard-ranking labels.

%% file: framework.tex
%------------------------------------------------------------------------
\begin{figure*}[!t]
	\begin{center}
		\includegraphics[width=0.9\linewidth]{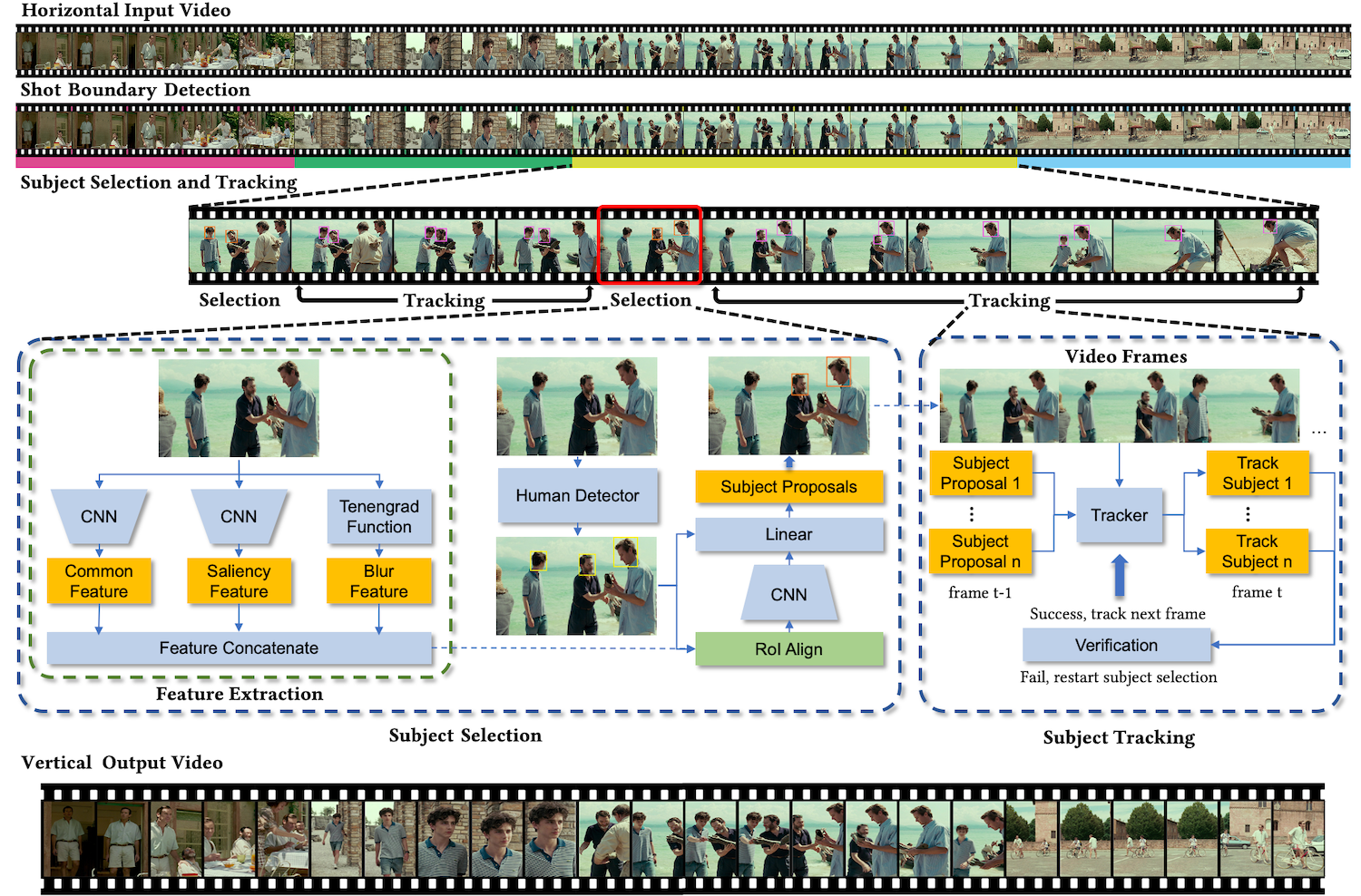}
	\end{center}
	\caption{Our proposed {\it H2V} framework (Section \ref{Framework Overview}). It consists of four steps: (1) Divide complete video into a series of footage. (2) Select subjects from keyframe (Section \ref{Subject Selection}), which is the first frame in the shot initially. (3) Tracking those subjects within consecutive frames until fails, then mark the current frame as a keyframe and return to step 2 iteratively. (4) Smooth the tracking trajectory and generate the vertical output video.}
	\setlength{\belowcaptionskip}{-0.3cm}   %调整图片标题与下文距离
	\label{fig:overview}
\end{figure*}

\section{{\it H2V} Framework}
\label{Framework Overview}
{\it H2V} video conversion clips horizontal video into a vertical format while keeping the most engaging content intact. Accordingly, one needs to identify and preserve the primary subject in every frame efficiently. To facilitate both the production-level accuracy and efficiency, our proposed {\it H2V} framework executes in a shot-based fashion. Firstly, a shot boundary detector TransNet~\cite{souvcek2019transnet} is employed to segment a horizontal input video into consecutive shots. Other video understanding methods~\cite{Liu2018TC3DTC,Tran2015LearningSF} for SDB could also be considered, which is beyond this paper's scope. Secondly, we apply our Rank-SS module to discover and select the primary subject within each shot. As the primary subject is mostly shot-stable, we finally bypass the frame-by-frame Rank-SS by tracking this subject throughout the shot with trajectory verification and smoothing.

\subsection{Subject Selection Criteria}
\label{sec:criteria}
Since human actors are primary subjects in most trending videos, {\it H2V} video conversion crops horizontal videos around the primary subject to reduce the loss of information and produce meaningful vertical content during the conversion process. To correctly identify the most primary human subject, we first discover all human objects in the scene using the DSFD face detector~\cite{li2019dsfd} and FreeAnchor body detector~\cite{zhang2019freeanchor}. Meanwhile, we prefer to utilizing a face detector since it is easier to maintain the completeness of a face than a body in the cropped area. Indeed, the ablation experiment in Section \ref{sec:img_a} proved that the face detector is more effective than a body detector. Then, selecting the primary subject from all is a highly empirical and subjective task, for which we summarize the following criteria under guidance from professional video editors:

\begin{itemize}
\item \textbf{The Central Criterion}: The primary subject tends to reside in the center of the scene.
\item \textbf{The Focal Criterion}: The primary subject appears within the focal length and free-from out-of-the-lens blurry. 
\item \textbf{The Proportional Criterion}: The primary subject tends to occupy the majority of the scene.
\item \textbf{The Postural Criterion}: The primary subject displays a more eye-catching posture rather than, for example, side-face or back-away. 
\item \textbf{The Stable Criterion}: (Video only) Primary subject usually shows no abrupt displacement within the same shot.
\end{itemize}

To fully practice the above criteria effectively and compactly, we design a subject selection component, as shown in Fig. \ref{fig:subselect}, which composes subject discovery, feature extraction, and our proposed {\it Sub-Select} module. The details are elaborated in Section \ref{Subject Selection}.

\subsection{Subject Tracking}
As each video shot usually focuses on the same primary subject, we refrain from the complicated frame-by-frame subject selection and track the primary subject selected as above throughout the shot. When professional editors performing {\it H2V} manually, video temporal smoothness is deliberately ensured with frame calibrating and interval smoothing. By mimicking this procedure, we design a subject tracking component integrating object-tracking based on SimaMask \cite{Wang2018FastOO}, verification, and temporal smoothness modules. Specifically, this component simultaneously tracks all subjects in the scene, and should a subject exit the scene, we re-verify the trajectory by triggering the subject selection component to re-initialize the primary subject. Whenever a subject disappears or is contaminated by similar distractors, the tracking confidence might be below the threshold. A subject exits the scene, which means that the vertical scene cannot cover all tracked subjects. All situations will trigger the verification module to restart the sub-select module for subject re-initialization. To further smooth the trajectory and suppress motion jitters, a Kalman Filter based motion model \cite{Zarchan2001FundamentalsOK} is incorporated.

%% file: subselect.tex
\begin{figure*}[!t]
	\begin{center}
		\includegraphics[width=1.\linewidth]{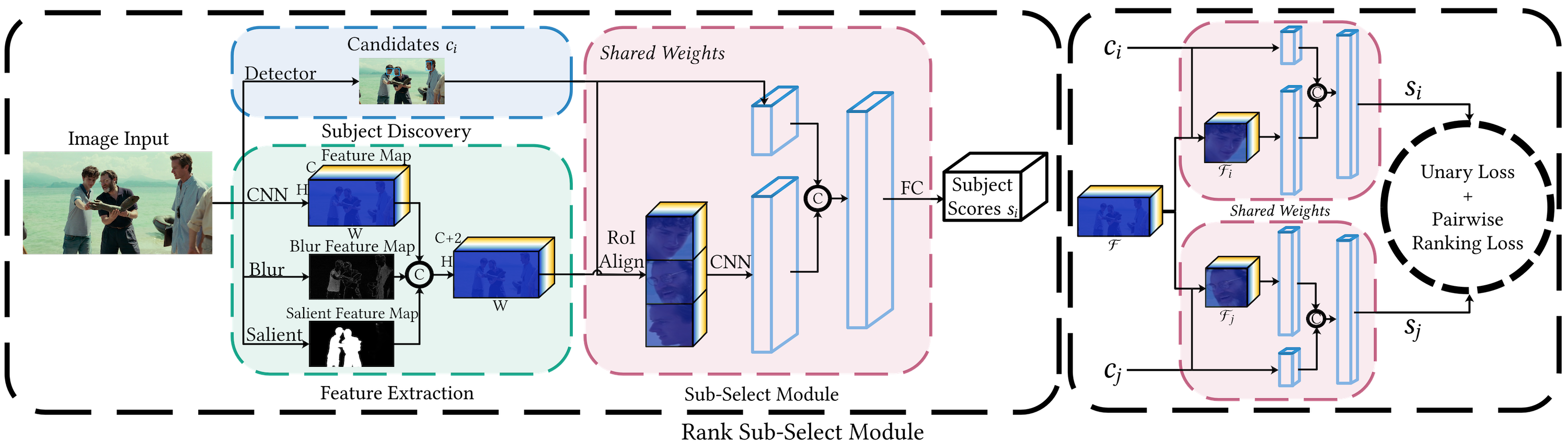}
	\end{center}
	\caption{An illustration of our Rank-SS module. Given an image, a detector is employed to generate face or body bounding boxes as subject candidates $\mathbf{c_i}$. Then, the RoIAlign~\cite{he2017mask} layer accepts the predicted bounding boxes and the concatenated features extracted by three feature extractors as input to generate fixed-size feature maps, which are fed into RCNN header. Finally, the vectorized feature is concatenated with the location priors of candidates and mapped to the subject prediction scores $\mathbf{s_i}$ by fully-connected layers.}
	\setlength{\belowcaptionskip}{-0.3cm}   %调整图片标题与下文距离
	\label{fig:subselect}
\end{figure*}

\section{Subject Selection}
\label{Subject Selection}
In this section we explain our subject selection component in-depth (shown in Fig. \ref{fig:subselect}), emphasizing feature extraction and the novel {\it Sub-Select} module. Particularly, the {\it Navie Sub-Select} (N-SS), {\it Deep Sub-Select} (D-SS), and {\it Rank Sub-Select} (Rank-SS) modules are discussed in turns to share more insights in solving subject selection.  

\subsection{Feature Extraction}
To fulfill the subject selection criteria described in Section \ref{sec:criteria}, we employ three different feature extraction strategies, as shown in Fig. \ref{fig:subselect}. For saliency feature extraction, we adopt the cascaded Partial Decoder model (CPD \cite{wu2019cascaded}) to generate a salient feature map $\mathcal{F}_{sal}$. Blur detection aims to detect Just Noticeable Blur (JNB) caused by defocusing that spans a small number of pixels in images. We utilize the traditional Thresholded Gradient Magnitude Maximization algorithm (Tenengrad \cite{tenenbaum1970accommodation,schlag1983implementation}) to extract the blur feature map $\mathcal{F}_b$. Additionally, we implement an ImageNet \cite{Krizhevsky2012ImageNetCW} pre-trained ResNet-50 \cite{he2016deep} to extract deep semantic embedding $\mathcal{F}_{e}$. In summary, the feature $\mathcal{F}$ exploited in our subject selection component is a concatenation as: $\mathcal{F} = [\mathcal{F}_{sal}, \mathcal{F}_{b}, \mathcal{F}_{e}].$

\subsection{N-SS and D-SS}
\label{sec:ndss}
Depending on the extracted feature, subject selection is a problem to compute the probability $s_{i}$ of each discovered human candidate $i$ for being the primary subject, and {\it H2V} conversion is then resolved by cropping the video around the most probable subject. The {\it Naive Sub-Select} (N-SS) module settles this problem by calculating the probability as the weighted summation of saliency and blur feature vectors. For each discovered subject candidate bounding box $c_i=[x_i,y_i,w_i,h_i]$ (where $i=1,2,...,N$), features extracted from within are abstracted via Global Average Pooling \cite{lin2013network} (GAP) to produce feature vector $f_i$:

\begin{equation}
    f_i = GAP(Crop([\mathcal{F}_{sal}, \mathcal{F}_{b}], c_i)).
\end{equation}
N-SS calculates the probability $s_{i}$ as: 

\begin{equation}
    s_i =  NSS(f_i,c_i; \mathbf{w}) = \mathbf{w}^\mathrm{T}[f_i,c_i],
\end{equation}
where we concatenate $f_i$ with $c_i$ to incorporate position and size information into consideration, conforming to the central and proportional criteria. $\mathbf{w}$ is the weights of the concatenated feature set manually. In the experiment, 0.3, 0.1, 0.3, 0.3 yields the best performance, each representing the weight for the saliency, blur, and bounding box size and position.

The main drawback of N-SS is the dependency on manually-set weights $\mathbf{w}$. As an improvement, the {\it Deep Sub-Select} (D-SS) module is designed with the same input as N-SS while adopting a Multi-Layer Perception (MLP) to learn the optimal weights in a data-driven way. This greatly enhances the capacity of {\it Sub-Select} module {\it w.r.t.} feature integration. Specifically, we implement three fully-connected layers followed by ReLU activation, and adopt Mean Squared Error (MSE) loss as the cost function to optimize D-SS, which is commonly used in the regression problem. The formulation of D-SS is: $s_i =  DSS(f_i,c_i; \mathbf{w}).$

\subsection{Rank-SS}
\label{sec:rankss}
Both N-SS and D-SS generate unary probability while overlooking the pairwise relationship among subject candidates. Concretely, prediction only considers the characteristics of the candidate itself, and the final predicted subject probability score is not related to others. In addition to larger absolute probability values, the primary subject is more distinctive from the non-primary ones {\it w.r.t.} higher probability ranking order. From this view, {\it Rank Sub-Select} (Rank-SS) module extends the D-SS module from regression into a ranking formulation, striving to select the primary subject more accurately with pairwise ranking supervision.

In addition to the salient feature, blur feature and bounding box size and position information as our selection basis, we design a RCNN-like \cite{girshick2015fast,ren2015faster} module (shown on the left-side in Fig. \ref{fig:subselect}) with deep semantic embedding. To better optimize the {\it Sub-Select} module in Rank-SS, we develop a new pairwise ranking-based supervision paradigm as illustrated on the right-side in Fig. \ref{fig:subselect}, and the Siamese architecture \cite{Li2018SiamRPNEO} has two identical {\it Sub-Select} module branches and is valid for pair-wise inputs. On top of a Siamese architecture, bounding boxes for subject $i$ and $j$ are simultaneously passed onto the Rank-SS module, together with the extracted feature $\mathcal{F}$ for the scene. Both branches in the Siamese architecture instantiate the same {\it Sub-Select} module, feature map $\mathcal{F}_{i}$ is pooled from bounding box $c_{i}$ on $\mathcal{F}$ with RoIAlign operation \cite{he2017mask}.  

\begin{equation}
    \mathcal{F}_i = RoIAlign(\mathcal{F}, c_i),
\end{equation}
where $\mathcal{F}_i$ is further vectorized through three cascaded ResNet \cite{he2016deep} bottleneck blocks (followed by GAP \cite{lin2013network}), then concatenated with the bounding box feature vector of $c_{i}$ and $c_{j}$ separately. The regression subject probability score $s_i$ is computed as: $s_i = RankSS(\mathcal{I}, c_i; \mathbf{w}),$
where $RankSS$ and $ \mathbf{w}$ indicate the Rank-SS network and its weights, respectively, $\mathcal{I}$ denotes the input image. To train the Rank-SS module, we implement both unary and pairwise loss functions. 

\textbf{Unary loss}
Point-wise Mean Squared Error (MSE) loss is adopted to measure the absolute difference between predicted and ground-truth probability score. The MSE loss for all $N$ candidates $c_i, i=1,2,...,N$ is calculated as:
\begin{equation}
    \mathcal{L}_{pt}(s) = \frac{1}{N} \sum_{i=1}^{N} (s_{i}-l_{i})^2 ,
\end{equation}
$l_{i}$ represents the ground-truth label of candidate $c_i$, which is 1 and 0 for subjects and non-subjects, respectively.

\textbf{Pairwise loss}
Our {\it H2V-142K} dataset is annotated with subject primality ranking labels, enabling pairwise supervision to improve the {\it Sub-Select} module in learning features and probabilities that better distinguish primary subject from the non-primary ones. Concretely, we adopt the margin-ranking loss \cite{Burges2005LearningTR} on $s_i$, $s_j$ generated from the Siamese Rank-SS module, and the associated ranking labels. To adjust Rank-SS output ranking orders compatible the annotation, we formulate:

\begin{equation}
\left\{
  \begin{aligned}
    \; s_i < s_j \quad&if\quad l_i < l_j \\
    \; s_i > s_j \quad&if\quad l_i > l_j
  \end{aligned}
\right..
\end{equation}
This pair-wise loss guide the Rank-SS ranking to the orientation of the given relative order, which is formulated as:
\begin{equation}
\begin{aligned}
\mathcal{L}_{pair}(s; \gamma) = &\frac{1}{N(N-1)} \sum_{i=1}^{N} \sum_{\substack{j=1 \\ j \neq i}}^{N}\mathop{\max}(0, (s_j - s_i) * \gamma_{(i,j)} + \epsilon),\\
&\left\{
  \begin{aligned}
    \; \gamma_{(i,j)} &= -1  &if\quad l_i < l_j \\
    \; \gamma_{(i,j)} &= 1   &if\quad l_i > l_j
  \end{aligned}
\right. ,
\end{aligned}
\end{equation}
where $\gamma$ is the rank label of the candidates pair. The margin $\epsilon$ controls the distance between $s_i$ and $s_j$. 

Subject probability scores $s$ of $N$ candidates are optimized with the combination of both unary and pairwise losses:
\begin{equation}
  \begin{aligned}
  \mathcal{L}=w_{pt} \mathcal{L}_{pt}(s) +w_{pair} \mathcal{L}_{pair}(s; \gamma),
  \end{aligned}
\end{equation}
where $w_{pt}$ and $w_{pair}$ denote the weights of $\mathcal{L}_{pt}$ and $\mathcal{L}_{pair}$.

%% file: experiment.tex
\section{Experiments} %\label{exp}

In this section, we first elaborately introduce ASR dataset, and evaluation metrics on both video and image data. On this dataset, we then present extensive experiments results {\it w.r.t.} our {\it H2V} framework, emphasizing the {\it Sub-Select} module with comprehensive subjective results.

\subsection{The ASR Dataset}

The ASR dataset \cite{siris2020inferring} is a large-scale salient object ranking dataset based on a combination of the
widely used MS-COCO dataset \cite{lin2014microsoft} with the SALICON dataset \cite{jiang2015salicon}. SALICON is built
on top of MS-COCO to provide mouse-trajectory-based fixations in addition to original objects' mask and bounding box annotations. The SALICON dataset provides two sources of fixation data: 1) fixation point sequences and 2) fixation maps for each image. The ASR dataset exploits these two sources to generate ground-truth saliency rank annotations. As the ASR dataset is not human-centered, we verify the generalization ability of the proposed method on this object-centered dataset.

\subsection{Evaluation Metrics}
To evaluate the performance on our {\it H2V-142K} dataset, we adopt the max Intersection-over-Union (max-IoU), min Central Distance Ratio (min-CDR), min Boundary Displacement Error (min-BDE), and mean Average Precision (mAP) metrics. Precisely, we measure the subject selection accuracy by calculating the max IoU over all ground-truth subject instances as: Since our new dataset has the images with multiple annotated subjects, we evaluate our subject selection result $c_y$ with each ground-truth subject by IoU. Then, we employ the max IoU to measure whether the subject is selected 
\begin{equation}
maxIoU = \max_i(\frac{Area(c_y \cap GT_i)}{Area(c_y \cup GT_i)}).
\end{equation}
where $c_y$ and $GT_i$ denote the bounding box of the predicted subject candidate and ground-truth subject instance $i$. Besides, min-CDR is employed to evaluate the precision of predicted bounding boxes as follows:

\begin{equation}
minCDR(y)=\min_i(\frac{\left \| y-\hat{y}_{i} \right \|}{w}),
\end{equation}
where $y$ and $\hat{y}_{i}$ indicate the center coordinates of predicted subject candidate and ground-truth subject instance $i$, the width of image $w$ is considered for normalization. Also, we adopt the same evaluation metric as~\cite{wang2018deep}, i.e., min-BDE to measure the accuracy of predicted subject. The min-BDE is defined as the average displacement of four edges between the predicted subject bounding box and the ground-truth rectangle:

\begin{equation}
    minBDE=\min_i(\frac{\sum_{j=1}^{4}(\left \| B_j-\hat{B}_{j}^{i} \right \|)}{4}),
\end{equation}
where $j \in \{left,right,up,bottom\}$, and $\{B_j\}_j$ denote the four edges of the predicted subject while $\{\hat{B}_{j}^{i}\}_j$ denote the four edges of ground-truth subject $i$.

In addition to the above image-oriented metrics, we also incorporate the average min-CDR (avg-min-CDR), Jitter Degree Ratio (JDR) and recall metrics to evaluate performance regarding videos. For average min-CDR, instead of setting $y$ as the center of bounding boxes, we set $y$ as the center of the cropped vertical frame and calculating the mean value over the whole video sequence. For JDR, we compute the sum of pair-frame pixel displacement {\it w.r.t.} the cropped center coordinates as below:
\begin{equation}
    JDR(\mathbf{y})= \sum_{t=1}^{T-1}(\frac{\left \| y_{t+1}-y_{t} \right \|_2}{w}),
\end{equation}
where $T$ indicates the total number of frames in the whole video sequence, and $w$ means the width of the frame. Recall metric \cite{Deselaers2008PanZS} refers to the percentage of the main subject that can be displayed on the clipping screen. Ideally, the main subject can be entirely displayed on the screen instead of being clipped out. The metric is described as follows:
\begin{equation}
    Recall=\sum_{t=1}^{T}(\max_i(\frac{Area(c_y \cap GT_i)}{Area(c_y)}))/T.
\end{equation}
where $c_y$ and $GT_i$ are the same meaning as in max-IoU.

\subsection{Experiments}
On the {\it H2V-142K} dataset, we conduct extensive experiments to evaluate the performance of our {\it Sub-Select} module and our {\it H2V} conversion framework. 

\subsubsection{{\it Sub-Select} Module}
We compare our ranking-based module with the state-of-the-art salient object detection CPD \cite{wu2019cascaded}, fixation prediction-based competitors \cite{he2019understanding}, image cropping \cite{Lu2019AnEN,zeng2020cropping}, as well as our naive and deep selection-based baselines, on the image subset of the {\it H2V-142K} dataset. For both SOD and FP methods, probability maps are generated for input images with pre-trained released models due to a lack of annotated data for our task. Then, the biggest contour in binarized probability maps is selected as the subject. The result position is represented as a bounding box and a centroid of the contour. As for image cropping, the traditional methods discard the irrelevant content and remain the enjoyable part of the image, but cannot output a fixed-size image. After transformation, the detected bounding box closest to the center of the cropped image is regarded as our selected subject.

\textbf{Implementation Details}. In our Rank-SS module and N-SS as well as D-SS baseline modules, we deploy DSFD~\cite{li2019dsfd} and FreeAnchor~\cite{zhang2019freeanchor} as the face and torso detectors. As for the integrated feature extraction described above, CPD~\cite{wu2019cascaded} is attached to produce the saliency detection response, and Tenengrad algorithm~\cite{tenenbaum1970accommodation,schlag1983implementation} is used to produce blur response in three proposed subject selection modules. In the Rank-SS module, an ImageNet pre-trained Resnet50 backbone is implemented to extract deep semantic embedding, and all feature maps are resized to stride 16 consistent with the embedding feature size. Input images are resized such that their shorter side is 600 pixels during training and testing. Regional feature size pooled by RoIAlign layer is 14$\times$14. 

For training modules, we employ SGD as an optimizer with an initial learning rate of 0.01 that decays 0.1 ratios per 10 epochs. The batch size of the input image is 4, and RoIs per image is increased to 20 by randomly perturbing subject candidates. To warm up the subject probability predictor, $w_{pair}$ is set as zero for the first 30 epoch. Then, the pairwise loss $\mathcal{L}_{pair}$ is added to train the Rank-SS module, which takes another 50 epochs to reach convergence. 

\textbf{Comparisons}.
In the following, N-SS and D-SS denote the naive and deep subject selection baselines, Rank-SS is our final ranking-based module. As shown in Table \ref{table:srss-image-c}, even our naive baseline with traditional features outperforms SOD~\cite{wu2019cascaded}, FP~\cite{he2019understanding} and image cropping~\cite{Lu2019AnEN,zeng2020cropping} competitors in all four metrics, achieved at least 0.86\% improvement in mAP. The deep regression baseline further improves upon N-SS by 19.81\% in max-IoU and 23.8\% in mAP, demonstrating the efficacy of deep features in the {\it H2V} conversion task. The soft label refers to the label information that does not contain the ranking order in the training data. We employ the correlation between the primary subject and the non-subject to construct a relative ranking order for training. The hard label refers to the annotated absolute ranking order between all human proposals. Our final ranking-based subject selection module reports the best overall selection accuracy with hard-label training data, surpassing D-SS by 4.65\%, exceeding the Rank-SS module with soft-label training data 2.24\%, and largely outperforming the SOD based model by 29.31\% in term of mAP. Also, our final ranking-based module reports the best min-CDR and min-BDE, which means that we achieve the best accuracy in predicting the subject bounding box. As for the speed tests, our naive baseline with traditional features reports the best FPS cause of naive liner feature combinations, and our ranking-based module maintains real-time processing performance while improving the accuracy.

\begin{table}[htb]
\renewcommand\tabcolsep{1.9pt} % 调整表格列间的长度
	\begin{center}
		\begin{tabular}{l|c c c c c}
        \hline
        Method & max-IoU $\uparrow$ & min-CDR $\downarrow$ & min-BDE $\downarrow$ & mAP $\uparrow$ & FPS $\uparrow$\\ 
        \hline
        SOD \cite{wu2019cascaded}& 59.59\% & 8.36\%&8.68\%& 65.17\%&12.7 \\
        FP \cite{he2019understanding}& 48.16\% & 9.78\%&9.26\%& 43.28\%& 15.6\\
        Image Cropping \cite{Lu2019AnEN}& 24.62\% & 37.87\%&24.55\%& 26.38\%& 21.2\\
        Image Cropping \cite{zeng2020cropping}& 15.79\% & 46.39\%&30.07\%& 17.24\%& 18.4\\
        \hline
        N-SS (Ours)&  62.47\% & 4.88\%&6.21\%& 66.03\%&\textbf{59.1}\\
        D-SS (Ours)& 82.28\% & 2.40\%&2.13\%& 89.83\%&31.3\\
        \hline
        Rank-SS/Soft (Ours)& 84.44\% & 1.55\%&1.98\%& 92.24\%&26.7 \\
        Rank-SS/Hard (Ours)& \textbf{92.37\%} & \textbf{1.01\%}&\textbf{1.89\%}& \textbf{94.48\%} &26.8\\
        \hline
		\end{tabular}
	\end{center}
	\caption{Comparison of our proposed {\it Sub-Select} modules with state-of-the-art Salient Object Detection, Fixation Prediction and Image Cropping methods in {\it H2V-142K} Image Subset. Please note that $\uparrow$ means the larger value representing the better result, while $\downarrow$ is the opposite.}
	\label{table:srss-image-c}
\end{table}

In addition to testing on {\it H2V-142K} Image Subset, we conduct experiments on the ASR dataset \cite{siris2020inferring} to evaluate the modules' generalization ability. As shown in Table \ref{table:srss-asr}, there are three kinds of training settings based on our training data, which include only {\it H2V-142K} Image Subset, only ASR dataset, and both two datasets. Our Rank-SS trained with {\it H2V-142K} Image Subset achieves 59.59\% in mAP, which significantly outperforms SOD~\cite{wu2019cascaded},  FP~\cite{he2019understanding} and image cropping \cite{Lu2019AnEN,zeng2020cropping} competitors. The Rank-SS trained with ASR dataset further improves upon the module trained with {\it H2V-142K} Image Subset by 5.72\% in max-IoU and 7.57\% in mAP, which benefits from the homogeneity of the dataset. The Rank-SS reports the best overall selection accuracy with both two datasets training data, surpassing the module trained with ASR dataset by 0.09\%, and largely outperforming the SOD based model by 19.77\% in term of mAP.

\begin{table}[htbp]
\renewcommand\tabcolsep{4.5pt} % 调整表格列间的长度
    \begin{tabular}{l|c c c c}
        \hline
        Method & max-IoU $\uparrow$ & min-CDR $\downarrow$ & min-BDE $\downarrow$ & mAP $\uparrow$\\ 
        \hline
        SOD \cite{wu2019cascaded}& 53.3\% & 11.94\% & 12.31\% & 47.48\% \\
        FP \cite{he2019understanding}& 23.29\% & 14.49\% & 19.04\% & 11.41\% \\
        Image Cropping \cite{Lu2019AnEN}& 12.19\% & 42.92\%&36.25\%& 10.42\%\\
        Image Cropping \cite{zeng2020cropping}& 17.84\% & 32.82\%&28.62\%& 27.22\%\\
        \hline
        Rank-SS ({\it H2V-142K})& 54.54\% & 4.69\% & 4.74\%& 59.59\%\\
        Rank-SS (ASR)& 60.26\% & 4.52\% & 4.22\% & 67.16\%  \\
        Rank-SS (ASR-ft) &  \textbf{60.42\%} & \textbf{4.39\%} & \textbf{4.13\%} & \textbf{67.25\%} \\
        \hline
    \end{tabular}
    \caption{Comparison of the proposed {\it Sub-Select} modules with state-of-the-art Salient Object Detection, Fixation Prediction and Image Cropping methods in the ASR dataset \cite{siris2020inferring}. }
    \label{table:srss-asr}
\end{table}

\textbf{Ablations}.
\label{sec:img_a}
To investigate each component's contribution in the Rank-SS module, we provide the ablation study results on our final ranking-based module in Table \ref{table:srss-image-a}. The last row in the table resides the complete module, and the first three rows show the results of ablating the blur detection (BD), saliency detection (SD), and positional feature (PF), respectively, where each of them shows the contribution of different degrees. Notably, the PF component demonstrates up to 15.69\% mAP decline upon ablation, proving that the spatial position and size are the most effective clue to subject selection. BD and SD contribute 2.93\% and 3.27\% in terms of mAP. The sixth row reveals that the face detector outperforms torso detector by 9.34\%, 1.92\%, 1.52\%, and 6.38\% in max-IoU, min-CDR, min-BDE, as well as mAP accordingly. This proves that the human face is more reliable evidence to support accurate and precise subject discovery. Finally, max-IoU and mAP decrease by 8.69\% and 3.1\% by ablating our margin-ranking loss (MRL), demonstrating that margin-ranking loss is especially valid in enforcing more spatially precise subject selection. The fifth row exposes that mean squared error loss (MSE) is equally indispensable in the Rank-SS module.

\begin{table}[htb]
\renewcommand\tabcolsep{4.0pt} % 调整表格列间的长度
	\begin{center}
		\begin{tabular}{l|c c c c}
		
        \hline
        Method & max-IoU $\uparrow$ & min-CDR $\downarrow$ & min-BDE $\downarrow$ & mAP $\uparrow$\\ 
        \hline
        Rank-SS (w/o BD)& 83.79\% & 1.74\%&2.02\%& 91.55\% \\
        Rank-SS (w/o SD)&  83.36\% & 1.75\%&2.09\%& 91.21\% \\
        Rank-SS (w/o SF) &  72.26\% & 5.26\%&7.34\%& 78.79\% \\
        \hline
        Rank-SS (w/o MRL)&  83.68\% & 1.72\%&2.08\%& 91.38\% \\
        Rank-SS (w/o MSE)&  83.74\% & 1.83\%&2.12\%& 91.55\% \\
        \hline
        Rank-SS (with body)& 83.03\% & 2.93\%&2.37\%& 88.10\% \\
        Rank-SS (with face)& \textbf{92.37\%} & \textbf{1.01\%}&\textbf{1.89\%}& \textbf{94.48\%} \\
        \hline
		\end{tabular}
	\end{center}
	\caption{Ablation study of our {\it Rank Sub-Select} module to investigate each component’s contribution. All ablation experiments take total 80 epochs for training. Using the face frame as the pre-selection result is better than using the body frame as the pre-selection.}
	\label{table:srss-image-a}
\end{table}

\begin{figure}[htbp]
\centering
\subfigure[max-IoU and mAP@max-IoU=0.5]{
\includegraphics[width=0.47\linewidth]{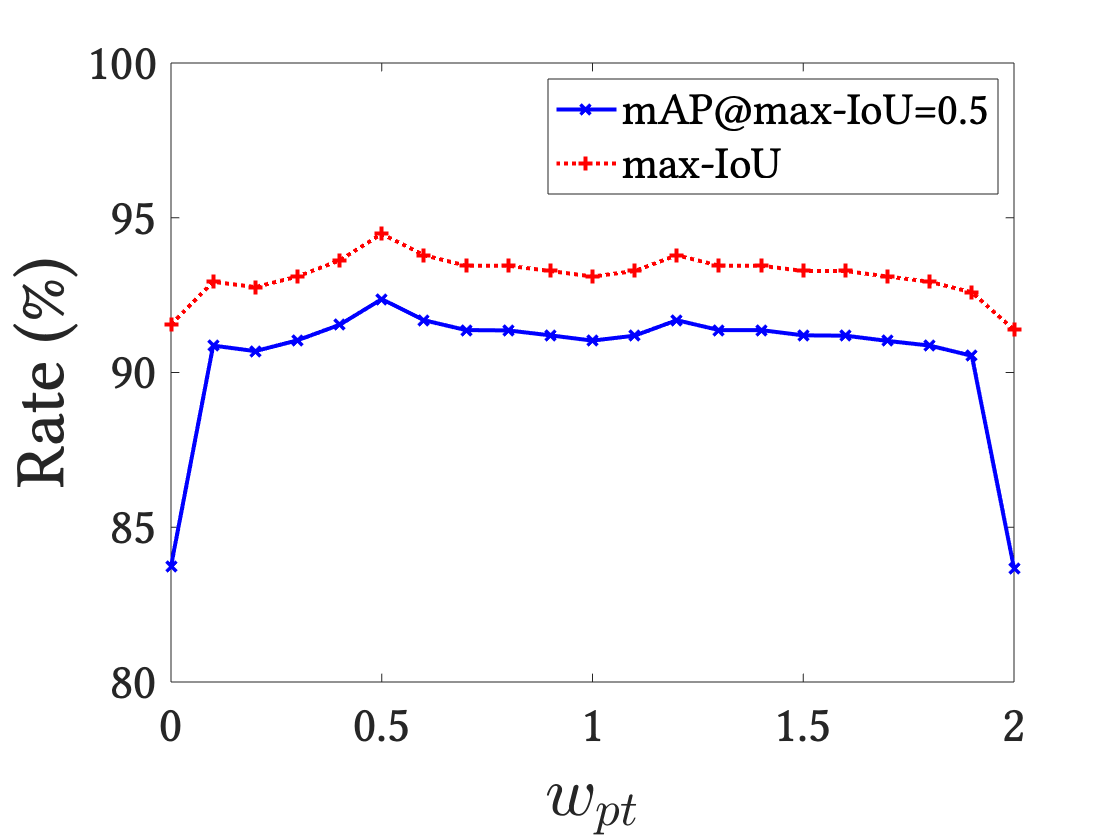}
}
\subfigure[min-CDR and min-BDE]{
\includegraphics[width=0.46\linewidth]{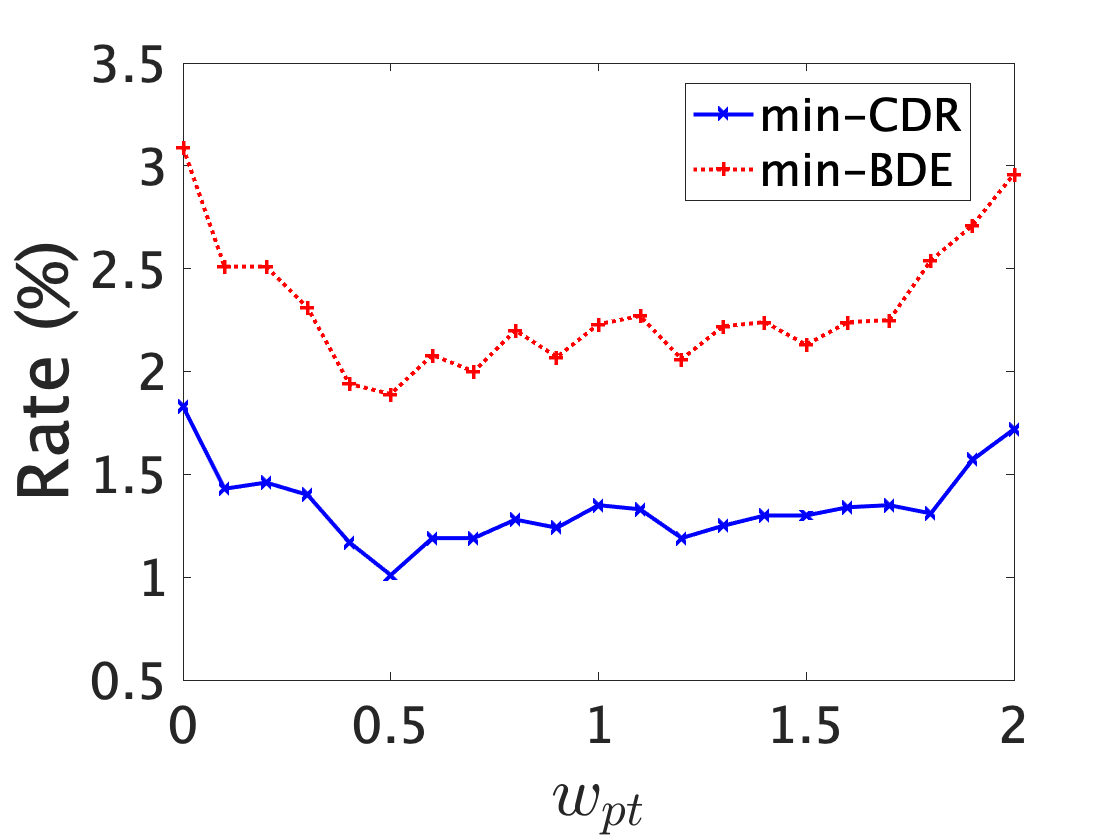}
}
\caption{Effect of the Number of Rank-SS weight $w_{pt}$ on the Image Subset of our {\it H2V-142K} Dataset.}
\label{fig:hyper-parameter}
\end{figure}

\textbf{Analysis of loss weights}. 
For our Rank-SS module, the weights of point-wise loss and pairwise loss are crucial hyper-parameters. Therefore, we further explore the effect of these two hyper-parameters by varying $w_{pt}$ from 0.0 to 2.0 s.t. $w_{pt}+w_{pair}=2$. Fig. \ref{fig:hyper-parameter} shows the impact of weights on the image subset of our {\it H2V-142K} dataset. From Table \ref{table:srss-image-c}, we note that the Rank-SS achieves the best performance while $w_{pt}$ is set to 0.5, and setting too large or too little number of $w_{pt}$ will affect the performance of subject selection. This is because a too large number of $w_{pt}$ cannot sufficiently utilize the ranking order of subject candidates and too little $w_{pt}$ will cause the prediction score to lose its meaning as the subject probability.

\subsubsection{{\it H2V} Framework}
We evaluate and compare the {\it H2V} framework on the video subset of our {\it H2V-142K} dataset, employing different {\it Sub-Select} variants, with other {\it H2V} frameworks based on SOD and FP subject selections, as well as video cropping framework~\cite{Frey2020Autoflip}. The video-based SOD anchor-diff~\cite{Yang2019AnchorDF} and FP Aclnet~\cite{wang2019revisiting} modules are executed on each frame to obtain the video results. The implementation is similar to the image-based SOD and FP methods.

\textbf{Comparisons}.
As shown in Table \ref{table:srss-video-c}, the {\it H2V} framework with the ranking-based {\it Sub-Select} module performs much better than SOD, FP and video cropping based approaches in both avg-min-CDR, JDR, and Recall metrics. As the FP approach is a point-based solution free from bounding boxes, it shows superior temporal stability and surpasses box-based SOD by 82.22\% and 22.82\% in terms of JDR and Recall. Although our Rank-SS based {\it H2V} is a region-based method, it still outperforms FP by 3.53\%, 1.48\% and 23.8\% improvements in all avg-min-CDR, JDR, and Recall metrics. This demonstrates the subject selection accuracy and temporal stability of our {\it H2V} framework.

\begin{table}[htb]
	\begin{center}
		\begin{tabular}{l|c c c c}
		\hline
        Method & avg-min-CDR $\downarrow$ & JDR $\downarrow$ & Recall $\uparrow$ & FPS $\uparrow$\\
        \hline
        Video SOD~\cite{Yang2019AnchorDF}& 14.49\%&  4.939& 22.91\%& 6.3\\
        Video FP~\cite{wang2019revisiting}&  14.47\%& 0.878&45.73\%& 15.6\\
        Video Cropping~\cite{Frey2020Autoflip}&  15.25\%& 5.382&20.09\%& 24.9\\
        \hline
        Our Framework&  \textbf{10.94\%}& \textbf{0.865}& \textbf{69.53\%} &\textbf{28.6}\\
        \hline
		\end{tabular}
	\end{center}
	\caption{Comparison of our {\it Rank Sub-Select} module based {\it H2V} framework with the state-of-the-art Video Salient Object Detection, Video Fixation Prediction and Video Cropping methods in {\it H2V-142K} Video Subset.}
	\label{table:srss-video-c}
\end{table}

\textbf{Ablations}.
Table \ref{table:srss-video-a} shows the ablation results of our {\it H2V} framework, where we investigate the contributions of the shot boundary detection (SBD) and subject tracking (ST) components. By ablating SBD, both avg-min-CDR, JDR and Recall drop significantly by up to 3.24\%, 61.74\%, and 15.88\%, which is induced by wrongful across-shot subject selection. Especially, the ground-truth subject varies from shot to shot, yet now the framework selects subject only once at the initial frame, thus rendering more errors. For ablating the ST component, on the other extreme, we execute {\it Sub-Select} at every frame, which causes FPS to decrease from 28.6 to 9.6. Consequently, avg-min-CDR increases by 6.18\% due to more accurate subject selection, while JDR and Recall decrease by 93.06\% and 12.8\% because of the absence of tracking-based temporal smoothness.

\begin{table}[htbp]
	\begin{center}
		\begin{tabular}{l|c c c c}
			\hline
        Method & avg-min-CDR $\downarrow$ & JDR $\downarrow$ & Recall $\uparrow$ & FPS $\uparrow$\\
        \hline
        %Our Framework (w/o SBD)&  14.18\%& 2.261& 53.65\%&\textbf{33.4}\\
        w/o SBD&  14.18\%& 2.261& 53.65\%&\textbf{33.4}\\
        w/o ST&  \textbf{4.76\%}& 12.468& \textbf{82.33\%}&9.6\\
        %Our Framework (w/o ST)&  \textbf{4.76\%}& 12.468& \textbf{82.33\%}&9.6\\
        \hline
        Our Framework & 10.94\%& \textbf{0.865}& 69.53\%&28.6\\
        \hline
		\end{tabular}
	\end{center}
	\setlength{\belowcaptionskip}{-0.3cm}   %调整图片标题与下文距离
	\caption{Ablation study of our {\it H2V} framework to investigate the contributions of shot boundary detection and subject tracking modules.}
	\label{table:srss-video-a}
\end{table}

\subsection{Subjective Evaluation}

\subsubsection{One-Way Repeated Measures ANOVA}
An independent sample (n=50) is recruited to complete the questionnaire, which contains 30 subjective evaluation questions consisting of one original image and three vertical results generated by SOD, FP, and Rank-SS, respectively. The participants are asked to report their evaluation of the three vertical images on Likert 5-point scale (1=bad, 5=excellent). To assess the difference in the performance of the three methods, we conduct one-way repeated measures ANOVA. The results are listed in Table \ref{table:anova}, which suggests that there is a significant difference in quality among the three methods ($F(2,98)=125, P<0.001,  \eta^2=0.487$). The {\it post-hoc} analyses (Table \ref{table:anova_pair}) reveal that the difference between Rank-SS and SOD is significant ($T=13.295, P<0.001$), suggesting that Rank-SS is better than SOD. Besides, there is a significant difference between Rank-SS and SOD ($T=11.015, P<0.001$), which demonstrates that Rank-SS also performs better than FP. However, no significant result is found between SOD and FP ($T=1.953, P=0.058$), indicating that the quality of cut result generated by SOD is similar to FP.

\begin{table}[htbp]
    \begin{center}
		\begin{tabular}{c|c c c|c|c}
        \hline
        \multirow{2}{*}{}& \multicolumn{3}{c|}{Method(Mean$\pm$SD)}& \multirow{2}{*}{$F$} & \multirow{2}{*}{$p$} \\ 
        \cline{2-4}
        &SOD\cite{wu2019cascaded}&FP\cite{he2019understanding}&Rank-SS & & \\
        \hline
        Score&2.63$\pm$0.67&2.76$\pm$0.70&4.09$\pm$0.67& 125& 0.000*** \\
        \hline
		\end{tabular}
	\end{center}
	\caption{The ANOVA experiment of {\it Rank Sub-Select} module with Salient Object Detection and Fixation Prediction methods.}
	\label{table:anova}
\end{table}

\begin{table}[htbp]
    \begin{center}
		\begin{tabular}{c|c|c|c|c|c}
        \hline
        \multirow{2}{*}{}& \multicolumn{3}{c|}{Paired Difference}& \multirow{2}{*}{$t$} & \multirow{2}{*}{$p$} \\ 
        \cline{2-4}
        &Mean&SD&SEM& & \\
        \hline
        FP - SOD&0.13&0.44&0.69& 1.953 & 0.058* \\
        RankSS - SOD&1.46&0.71&0.11& 13.295 & 0.000*** \\
        RankSS - FP&1.33&0.78&0.12& 11.015 & 0.000*** \\
        \hline
		\end{tabular}
	\end{center}
    \caption{The {\it post-hoc} analyses of ANOVA experiment.}
    \label{table:anova_pair}
\end{table}

\subsubsection{User-Feedback Results}
In applications, {\it H2V} video conversion is a more user-oriented task whose performance is better evaluated by user-feedback. After deploying our {\it H2V} framework onto Youku video-sharing website, we have collected rich user-feedback data generated from converted films, TV series, and variety shows, totally covering more than 100 Occupationally-Generated Content (OGC) with 10 million views. Specifically, the audit pass rate of converted images and videos is 98\% and 94\%, respectively. Image-wise, vertical video cover image converted by {\it H2V} gains video exposure up to 1.5 million times per day. Video-wise, efficiency indexes such as the Click Through Rate (CTR) and Bounce Rate (BR) of vertical videos converted by {\it H2V} is on par with the ones produced manually. In addition to online data explained as above, we invite a group of professional video practitioners to participate in an offline survey, wherein videos converted by {\it H2V} reports a 3\% bad case rate, way past the 5-10\% available rate.

\subsubsection{Strength and Weakness}

\begin{figure}[htbp]
  \centering
  \includegraphics[width=\linewidth]{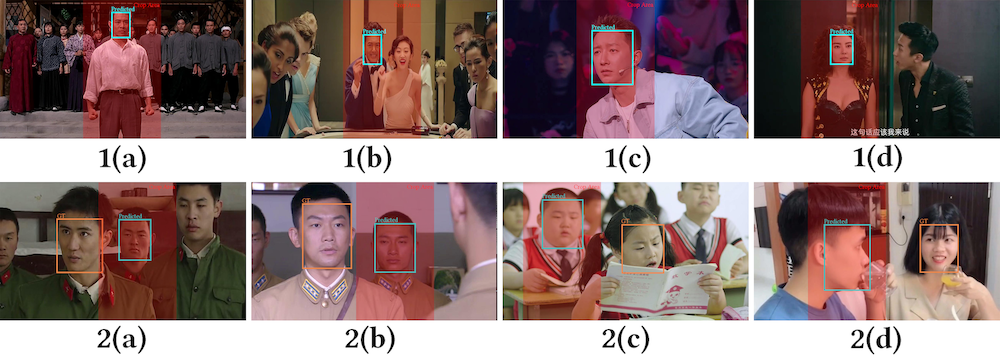}
  \caption{Visualization of results from {\it Sub-Select} Image Dataset. The blue and yellow bounding boxes are the selected subjects and ground-truth subjects, respectively. The final cropping area is highlighted as red. The first row denotes correct instances, and the second row presents bad cases.}
  \label{fig:case}
\end{figure}

\begin{figure*}[!ht]
  \centering
  \includegraphics[width=0.9\linewidth]{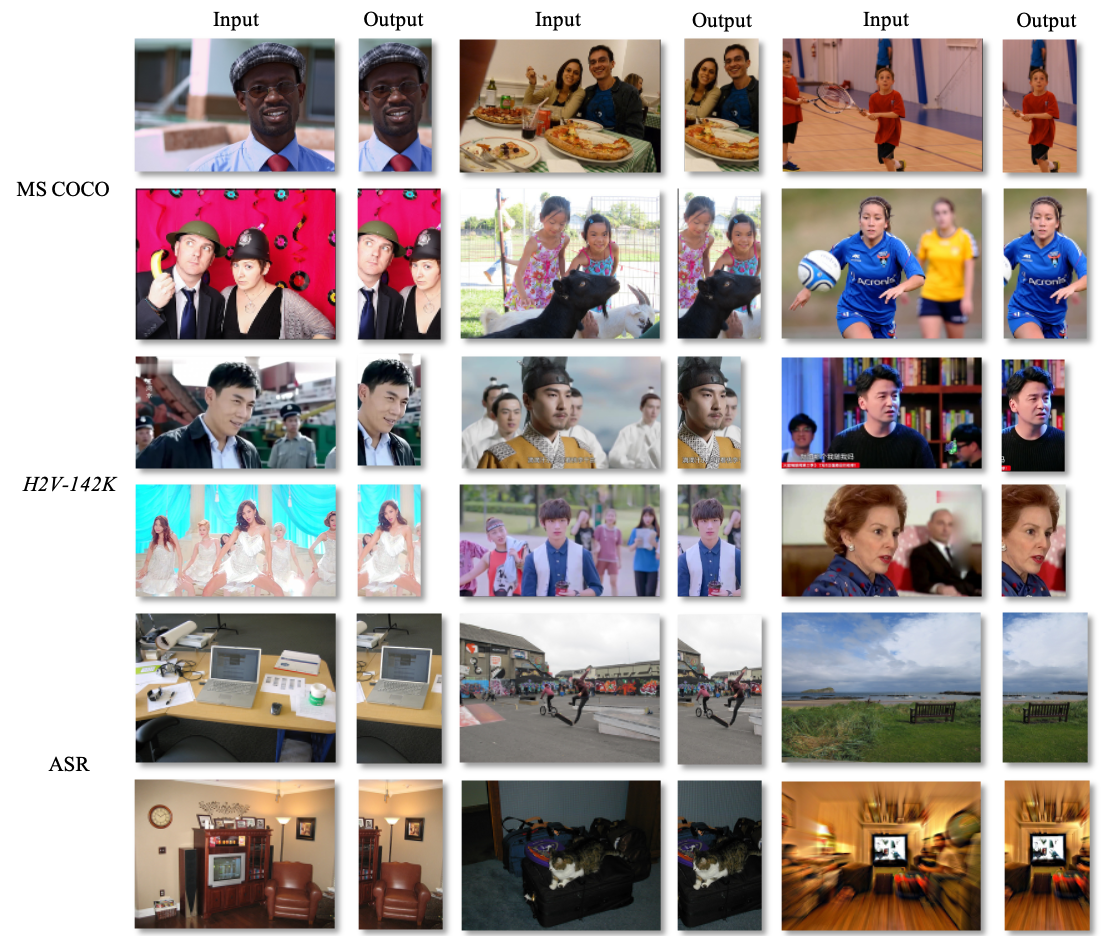}
  \caption{Visualization of results in the MSCOCO 2017 Val dataset~\cite{Lin2014MicrosoftCC}, proposed H2V-142K dataset, and ASR dataset \cite{siris2020inferring}. The original input and final 9:16 frame output are on either side of the picture. It worth noticing that the horizontal target frame can be a different value such as 1:1, 3:4, {\it etc}. As the primary subject selected, our {\it H2V} framework is able to generate multiply outputs with different sizes simultaneously.}
  \label{fig:coco}
\end{figure*}

As illustrated in the first row of Fig. \ref{fig:case}, our {\it H2V} framework successfully selects the primary subject from background distractors (1a and 1c). It can also discard the pseudo-subject, who is not facing the camera directly (1d). Moreover, {\it H2V} can incorporate human closely-located with the selected primary subject. More visualization of results in the MSCOCO 2017 Val dataset~\cite{Lin2014MicrosoftCC}, proposed {\it H2V-142K} dataset, and ASR dataset \cite{siris2020inferring} is shown in Fig. \ref{fig:coco}. As it can be seen that our method generalizes well on common objects not limited to humans. 

The second row of Fig. \ref{fig:case} shows several bad cases of our framework. By large, bad cases occur when the criteria described in Section \ref{sec:criteria} contradict each other. In 2a and 2b, the central criterion overwhelms the proportional criterion, and 2d demonstrates the case when the proportional criterion overwhelms the postural criterion. Sub-figure 2c depicts the missed detection. 

Finally, the limitations of our {\it H2V} framework include: 1) it is object-oriented method that cannot process scenery lens in documentary or contents with rarely-seen subjects such as plants that is not able to be detected by object detectors. 2) our framework may fail in strongly dynamic shots where the camera moves like Dolly, Truck and Zoom are utilized, resulting in jittering. 3) the {\it H2V-142K} dataset is a subset of {\it H2V} task containing human subjects only. It is still worthy of exploring more general cases in future works.

%% file: conclusion.tex
\section{Conclusions}
% In this paper, we pioneer the first effort in resolving automated horizontal-to-vertical ({\it H2V}) video conversion, which is a highly challenging and rewarding task in the modern video industry. Specifically, we propose the {\it H2V} framework, which for the first time defines the H2V pipeline and provides a fully-automated framework. Among all the well-performing modules in {\it H2V}, we highlight our design of the {\it Sub-Select} module, which effectively discovers and selects the primary subject in the scene via multi-cue feature integration and region-based object ranking. To build and evaluate {\it H2V}, we also collect and publicize a large-scale {\it H2V-142K} dataset, wherein 132k frames in 125 videos and 9,500 images are carefully annotated with subject building boxes by human-annotators. Extensive experiments with {\it H2V} are conducted on top of the {\it H2V-142K} dataset, where vast user-feedback data and metrics strongly reveal the efficacy and superiority of our {\it H2V} framework. Upon the completion of the paper, our {\it H2V} framework has been deployed online with massive throughput, and we hope this paper can pave the way for more successful endeavors to come. 

In this work, we introduce the first fully automatic and commercialized horizontal-to-vertical video conversion solution, which tackles subject-preserving clipping by integrating shot detection, subject selection, object tracking, and video cropping. Among all well-performing components, we highlight the {\it Sub-Select} module which effectively discovers and selects the primary subject via multi-cue feature integration and region-based object ranking. For the development of horizontal-to-vertical solutions, we hereby make the large-scale {\it H2V-142K} dataset publicly available, wherein 132K frames in 125 videos and 9,500 images are carefully annotated with primary subject building boxes. Extensive experiments with {\it H2V} are conducted on {\it H2V-142K} and related object detection datasets, where both accuracy metrics and vast user-feedback data reveal the efficacy and superiority of our {\it H2V} framework. Upon the completion of this paper, our {\it H2V} framework has been successfully deployed online hosting massive throughput, and we hope this paper can pave the way for more successful endeavors.